%% file: nunes2024cvpr.tex
\documentclass[10pt,twocolumn,letterpaper]{article}

\usepackage[camera-ready]{cvpr}      %
\usepackage{float}
\usepackage{amsmath}
\renewcommand{\and}{\hspace{0.6cm}}

\input{preamble}

\definecolor{cvprblue}{rgb}{0.21,0.49,0.74}
\usepackage[pagebackref,breaklinks,colorlinks,citecolor=cvprblue]{hyperref}

\def\paperID{2112} %
\def\confName{CVPR}
\def\confYear{2024}

\title{Scaling Diffusion Models to Real-World 3D LiDAR Scene Completion\vspace{-0.3cm}}

\author{
Lucas Nunes$^1$
\and
Rodrigo Marcuzzi$^1$
\and
Benedikt Mersch$^1$\\
Jens Behley$^1$
\and
Cyrill Stachniss$^{1,2}$\\
{\small $^1$Center for Robotics, University of Bonn \and $^2$Lamarr Institute for Machine Learning and Artificial Intelligence}\\
{\tt\small \{firstname.lastname\}@igg.uni-bonn.de}
}\vspace{-0.1cm}

\begin{document}
\twocolumn[{
\renewcommand\twocolumn[1][]{#1}%
\maketitle
\centering
\vspace{-0.5cm}
\includegraphics[width=1.0\textwidth]{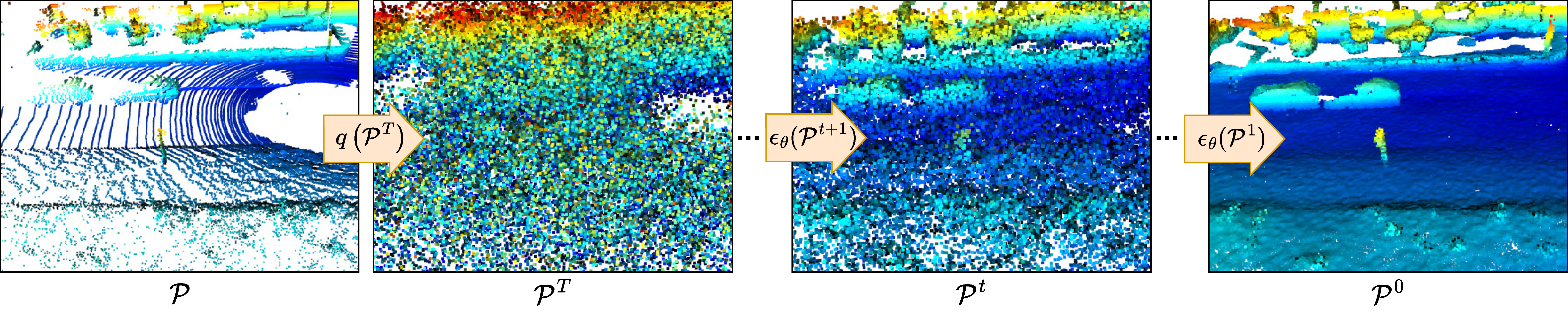}
\vspace{-0.5cm}
\captionof{figure}{Starting from a single input scan $\mathcal{P}$, we add Gaussian noise to each point, defining the noisy input $\mathcal{P}^T$. Then, we use our trained noise predictor~$\epsilon_\theta$ to denoise $\mathcal{P}^T$
iteratively until arriving at $\mathcal{P}^0$, yielding a completed representation of the 3D scene. \vspace{2em}}
\vspace{-0.3cm}
\label{fig:motivation}
}]

\begin{abstract}
    Computer vision techniques play a central role in the perception stack of autonomous vehicles. Such methods are employed to perceive the
    vehicle surroundings given sensor data. 3D LiDAR sensors are commonly used to collect sparse 3D point clouds from
    the scene. However, compared to human perception, such systems struggle to deduce the unseen parts of the scene given those sparse point clouds. In this matter, the scene
    completion task aims at predicting the gaps in the LiDAR measurements to achieve a more complete scene representation. Given the
    promising results of recent diffusion models as generative models for images, we propose extending them to achieve scene completion from a single 3D LiDAR scan.
    Previous works used diffusion models over range images extracted from LiDAR data, directly applying image-based diffusion methods. Distinctly,
    we propose to directly operate on the points, reformulating the noising and denoising diffusion process such that it can efficiently work at scene scale.
    Together with our approach, we propose a regularization loss to stabilize the noise predicted during the denoising process.
    Our experimental evaluation shows that our method can complete the scene given a single LiDAR scan as input, producing a scene with more details compared
    to state-of-the-art scene completion methods. We believe that our proposed diffusion process formulation can support further research in diffusion
	models applied to scene-scale point cloud data. \footnote{Code: \url{https://github.com/PRBonn/LiDiff}}
\end{abstract}

\section{Introduction}
\label{sec:intro}
\vspace{0.2cm}

Perception systems are a crucial component of self-driving cars, enabling them to understand their surroundings and safely navigate through it. Such systems
rely on the data collected by the sensors installed on the vehicle to perceive the environment but fail to deduce areas only partially observable by the sensor.
For a human it is comparably rather simple to infer the complete scene from the scene context. Especially in autonomous driving, LiDAR sensors are employed to collect 3D
information of the vehicle surroundings to enable safe navigation. Despite the accuracy of those sensors, collected point clouds are sparse, with large gaps between the data points
measured by the sensor beams. Being able to complete the measured scene can add valuable information to perception systems, helping
to improve different tasks such as object detection~\cite{xiong2023cvpr-lcrf}, localization~\cite{vizzo2022ral} or navigation~\cite{popovic2021ral}.

Scene completion tries to infer the missing parts of a scene, providing a dense and more complete scene representation. Given the LiDAR data sparsity,
having a way to fill the gaps of non-observed regions is helpful to enlarge the incomplete data measured by the sensor. Previously, this task was tackled using
paired RGB images and LiDAR point clouds by inferring depth maps from an RGB image supervised by the LiDAR depth measurements~\cite{xu2019iccv-dcfs,ma2018icra,ma2019icra-sssdcf,fu2020iros-dcvi}.
Other approaches~\cite{murez2020eccv,vizzo2022ral,li2023icra-llce} employ signed distance fields~(SDF) where the scene is represented as a voxel grid where each voxel stores its distance to the closest
surface in the point cloud. Such methods approximate the scene by a surface representation, losing details
usually present in real-world data since these approaches are limited to the voxel resolution. As an extension to this task, semantic scene completion has emerged~\cite{roldao2020threedv,rist2021pmai,li2023icra-llce},
where the goal is to infer an occupancy voxel grid with a semantic label associated to each voxel. However, those methods require large amounts of labeled data and
operate at a predefined fixed voxel grid resolution. More recently, denoising diffusion probabilistic models (DDPM) were employed in the context of self-driving cars~\cite{zyrianov2022eccv,lee2023arxiv,nakashima2023arxiv}
relying on image representations of the LiDAR data, such as range images~\cite{zyrianov2022eccv, nakashima2023arxiv} or a discrete diffusion process formulation, inferring the occupancy on
a predefined voxel grid~\cite{lee2023arxiv}.

In this work, we propose a diffusion scheme for 3D data operating at point level and at scene scale. We exploit the generative properties of DDPMs to infer the
unseen regions of a scene measured by a 3D LiDAR sensor, achieving scene completion from a single point cloud as illustrated in \cref{fig:motivation}. We reformulate the
(de)noising scheme used in DDPMs by adding noise locally to each point without scaling the input data to the noise range, allowing the model to learn
detailed structural information of the scene. Furthermore, we propose a regularization to stabilize the DDPMs during training, approximating the predicted
noise distribution closer to the real data. We compare our method with different scene completion approaches and conduct extensive experiments to validate
our proposed scene-scale 3D diffusion scheme. In summary, our key contributions are:

\begin{itemize}
	\item We propose a novel scene-scale diffusion scheme for 3D sensor data that operates at the point level.
	\item We propose a regularization that approximates the predicted noise to the expected noise distribution.
	\item Our method can generate more fine-grained details compared to previous methods.
	\item Our approach achieves competitive performance in scene completion compared to previous diffusion and non-diffusion methods.
\end{itemize}

\section{Related Work}
\label{sec:related_work}

\textbf{Scene completion} aims at inferring missing 3D scene information given an incomplete sensor measurement. This inference
of unseen information can be helpful for perception tasks~\cite{xiong2023cvpr-lcrf}, localization~\cite{vizzo2022ral} or navigation~\cite{popovic2021ral}. Some works~\cite{xu2019iccv-dcfs,ma2018icra,ma2019icra-sssdcf,fu2020iros-dcvi}
tackled this task by jointly extracting information from paired RBG images and LiDAR point clouds, predicting a depth map from an RGB image supervised by the LiDAR data.
Differently, other methods~\cite{vizzo2022ral,li2023icra-llce} approach the problem by optimizing a signed distance field (SDF) given only the LiDAR measurements, representing the scene as a voxel grid
where each voxel stores its distance to the closest surface in the scene. However, such methods are bound to the
voxel resolution and lose details in the scene due to the discretization by voxels. Distinctly, our approach works directly on the points and exploits the
generative properties of DDPMs to complete the unseen data without relying on a voxel grid representation.

\textbf{Semantic scene completion} has been of great interest more recently due to the availability of large datasets with semantic labels~\cite{geiger2012cvpr,behley2019iccv,caesar2020cvpr,sun2020cvpr,behley2021ijrr,fong2022icra,liao2022pami}.
This task extends the scene completion task by predicting a semantic label for each occupied voxel~\cite{roldao2020threedv,rist2021pmai,li2023icra-llce}. However, those methods are also tightly bound to the
voxel grid resolution, which usually has a low resolution due to memory limitations. Besides operating at point level, given the
recent research effort for DDPMs, our method could also later be extended to predict a semantic class for each generated point.

\textbf{Denoising diffusion probabilistic models} have gained attention due to their high-quality results in image generation~\cite{ho2020neurips,dhariwal2021neurips,nichol2021pmlr,ramesh2021pmlr,rombach2022cvpr,zhou2022cvpr-tltf,zhou2023cvpr-sdft,peebles2023iccv}.
Besides that, conditioned diffusion models gained even more relevance due to the possibility of generating data towards an input condition~\cite{ho2021neuripsws,balaji2022arxiv,zhang2023iccv}. The drawback of DDPMs is usually the time needed during the denoising process. For that reason,
many efforts have been put to achieve a faster generation, \eg, by doing a distillation of the denoising model~\cite{salimans2022iclr,meng2023cvpr} or by
analytically approximating the denoising steps solution to reduce the amount of steps needed~\cite{song2021iclr,karras2022neurips,lu2022neurips,lu2023arxiv}.

\textbf{Diffusion models for 3D data} have been investigated due to their promising performance in the image domain. Such methods~\cite{zhou2021iccv-sgac,luo2021cvpr,lyu2022iclr,zeng2022neurips,sanghi2022cvpr,sanghi2023cvpr,xu2023cvpr-dzts}
are focused on single object shapes, achieving novel object shape generation or completion. Few works~\cite{zyrianov2022eccv,lee2023arxiv,nakashima2023arxiv} target real-world data generation.
Some works~\cite{zyrianov2022eccv,nakashima2023arxiv} rely on projecting the 3D data to an image-based representation such as range images, such that the methods
proposed in the image domain can be directly applied. For such approaches, the 3D scene cannot be completed since when reprojecting the image
to the 3D world, some regions do not have any information due to occlusions in the projected point cloud. Lee \etal~\cite{lee2023arxiv} achieves scene-scale 3D data generation using a discrete
diffusion model formulation and a fixed voxel grid representation of the environment. The model is then used to infer for each voxel whether it is occupied,
and a semantic label is predicted. Different from previous works, our method operates directly at point level and does not rely on a grid
representation or projection to the image domain.

Given the recent advances in DDPMs for data generation, we propose a formulation of the denoising diffusion process that
works at point level, achieving competitive performance in scene-scale diffusion scene completion. Our formulation enables the use of DDPMs to generate
scene scale, real-world-like data without relying on any discretization or projection of the LiDAR data.

\section{Approach}

We propose using DDPMs to achieve scene completion from a single 3D LiDAR scan as input.
First, we reformulate the DDPMs~\cite{zhou2021iccv-sgac,luo2021cvpr,lyu2022iclr} to work at scene scale. Instead of normalizing the input point cloud,
we add and predict the noise locally for each point. During the denoising process, we condition the noise
prediction with the input scan such that the final scene retains the structural information from the
input scan while inferring the missing parts. In this formulation, the initial point cloud
is a noisy version of the input scan and the networks task is to denoise it to get the complete scene as depicted in \cref{fig:motivation}.
Next, we provide the needed background on diffusion models and describe the individual components of our approach. 

\subsection{Denoising diffusion probabilistic models}
\label{subsec:ddpm}

Denoising diffusion probabilistic models~\cite{ho2020neurips,dhariwal2021neurips,nichol2021pmlr} formulate the data generation as an iterative denoising process.
Commonly, the model starts from Gaussian noise~\cite{ho2020neurips,dhariwal2021neurips,nichol2021pmlr} and iteratively removes noise from the
input until it converges to the target output (\eg, images~\cite{ho2020neurips,dhariwal2021neurips,nichol2021pmlr,ramesh2021pmlr,rombach2022cvpr,zhou2022cvpr-tltf,zhou2023cvpr-sdft,peebles2023iccv} or shapes~\cite{zhou2021iccv-sgac,luo2021cvpr,lyu2022iclr,zeng2022neurips,sanghi2022cvpr,sanghi2023cvpr,xu2023cvpr-dzts}). This can be achieved by defining a
forward diffusion process where noise is iteratively added $T$ times to the target data. Then, the model is trained
to predict the noise added at each step~$t$. By predicting the noise at each step~$t$ and removing it, the denoised sample
should be closer to the target training data.

\textbf{The diffusion process} as formulated by Ho \etal~\cite{ho2020neurips} can be generally written as follows. Given a sample ${\boldsymbol{x}^0 \sim q\hspace{-0.075cm}\left(\boldsymbol{x}\right)}$ from a target data distribution,
the diffusion process adds noise to $\boldsymbol{x}^0$ over $T$ steps, resulting in $\boldsymbol{x}^1,\dots,\boldsymbol{x}^{T}$, where ${q\hspace{-0.075cm}\left(\boldsymbol{x}^{T}\right) \approx \mathcal{N}\hspace{-0.075cm}\left(\boldsymbol{0},\boldsymbol{I}\right)}$,
where ${\mathcal{N}\hspace{-0.075cm}\left(\boldsymbol{0},\boldsymbol{I}\right)}$ is a normal distribution with mean $\boldsymbol{0}$ and the identity matrix $\boldsymbol{I}$ as diagonal covariance.
This diffusion process is parameterized by a sequence of defined noise factors $\beta_1,\dots,\beta_T$, where iteratively at each step $t$, Gaussian noise is sampled and
added to~$\boldsymbol{x}^{t-1}$ given $\beta_t$. This can be simplified to sample $\boldsymbol{x}^{t}$ from $\boldsymbol{x}^{0}$, without computing the intermediary steps
$\boldsymbol{x}^1,\dots,\boldsymbol{x}^{t-1}$. To do so, Ho \etal~\cite{ho2020neurips} define $\alpha_t = 1 - \beta_t$ and
$\overline{\alpha}_t = \prod_{i=1}^{t} \alpha_i$, and $\boldsymbol{x}^{t}$ can be sampled as:
\begin{equation}
    \label{eq:closed_form}
    \boldsymbol{x}^t = \sqrt{\overline{\alpha}_t}\boldsymbol{x}^0 + \sqrt{1 - \overline{\alpha}_t}\boldsymbol{\epsilon},
\end{equation}

\noindent where ${\boldsymbol{\epsilon} \sim \mathcal{N}\hspace{-0.075cm}\left(\boldsymbol{0},\boldsymbol{I}\right)}$. Note that when $T$ is large enough
${q\hspace{-0.075cm}\left(\boldsymbol{x}^T\right) \approx \mathcal{N}\hspace{-0.075cm}\left(\boldsymbol{0},\boldsymbol{I}\right)}$, since $\overline{\alpha}_T$ gets closer to
zero.

\textbf{The denoising process} aims to undo the $T$ noising steps by predicting the noise $\boldsymbol{\epsilon}$ added at each step $t$~\cite{ho2020neurips}.
Given an initial $\boldsymbol{x}^T$, we want to reverse the diffusion process and get to $\boldsymbol{x}^0$. The reverse
diffusion step can be written as:
\begin{equation}
    \small
    \label{eq:reverse_diff}
    \boldsymbol{x}^{t-1} = \boldsymbol{x}^t - \frac{1 - \alpha_t}{\sqrt{1 - \overline{\alpha}_t}}\boldsymbol{\epsilon}_\theta\hspace{-0.075cm}\left(\boldsymbol{x}^t,t\right) + \frac{1-\overline{\alpha}_{t-1}}{1-\overline{\alpha}_t}\beta_t\mathcal{N}\hspace{-0.075cm}\left(\boldsymbol{0},\boldsymbol{I}\right),
\end{equation}

\noindent where $\boldsymbol{\epsilon}_\theta(\boldsymbol{x}^t,t)$ is the noise predicted from $\boldsymbol{x}^t$ at step $t$.

This generation can also be guided given a condition~$\boldsymbol{c}$. This conditional generation can either stem from a pre-trained encoder~\cite{dhariwal2021neurips}
or from classifier-free guidance~\cite{ho2021neuripsws}, where the encoder is trained together with the noise predictor. In our case, we use the
classifier-free guidance since it does not require a pre-trained encoder. With the classifier-free guidance, the model is trained to learn the conditional and unconditional
noise distribution. In this case, at each training step the model has a probability $p$ of predicting the unconditional noise distribution,
where the conditioning is set to a null token, \ie, ${\boldsymbol{c}=\emptyset}$.

\textbf{The training process} optimizes the denoising model to predict the noise $\boldsymbol{\epsilon}$ added at step $t$ to a given input.
Given an input $\boldsymbol{x}^0$ and a condition $\boldsymbol{c}$, a random step $t \in [0,T]$ is sampled, and $\boldsymbol{x}^t$ is
sampled from \cref{eq:closed_form} with a Gaussian noise $\boldsymbol{\epsilon}$. Then, from $\boldsymbol{x}^t$, $\boldsymbol{c}$ and $t$, the model computes the
noise prediction, supervising it with an $\mathcal{L}_2$ loss:\vspace{-0.06cm}
\begin{equation}
    \label{eq:diff_loss}
    \mathcal{L}\hspace{-0.075cm}\left(\boldsymbol{x}^t, \tilde{\boldsymbol{c}}, t\right) = \big\lVert \boldsymbol{\epsilon} - \boldsymbol{\epsilon}_\theta\hspace{-0.075cm}\left(\boldsymbol{x}^t, \tilde{\boldsymbol{c}}, t\right) \big\rVert ^2,
\vspace{-0.06cm}
\end{equation}

\noindent with ${\tilde{\boldsymbol{c}} \sim \mathcal{B}\hspace{-0.075cm}\left(p\right)}$ where $\mathcal{B}$ is a Bernoulli distribution with outcomes ${\{\emptyset, \boldsymbol{c}\}}$ with probability $p$ that $\emptyset$ occurs.

\textbf{The inference} starts from an initial ${\boldsymbol{x}^T \sim \mathcal{N}\hspace{-0.075cm}\left(\boldsymbol{0},\boldsymbol{I}\right)}$ and iteratively denoise it to get~$\boldsymbol{x}^0$.
For the classifier-free guidance~\cite{ho2021neuripsws}, we predict the conditional and unconditional noise distribution
and compute the final predicted noise as:
\begin{equation}
    \label{eq:class_free}
    \small
    \boldsymbol{\epsilon}'_\theta\hspace{-0.075cm}\left(\boldsymbol{x}^t,\boldsymbol{c},t\right) = \boldsymbol{\epsilon}_\theta\hspace{-0.075cm}\left(\boldsymbol{x}^t,\emptyset,t\right) + s\hspace{-0.075cm}\left[\boldsymbol{\epsilon}_\theta\hspace{-0.075cm}\left(\boldsymbol{x}^t,\boldsymbol{c},t\right) - \boldsymbol{\epsilon}_\theta\hspace{-0.075cm}\left(\boldsymbol{x}^t,\emptyset,t\right)\right],
\end{equation}

\noindent where $s \in \mathbb{R}$ is a parameter that weights the conditioning to~$\boldsymbol{c}$, and $\boldsymbol{\epsilon}_\theta\hspace{-0.075cm}\left(\boldsymbol{x}^t,\emptyset,t\right)$ is the unconditional noise
prediction.

With \cref{eq:class_free} we can compute the noise at any step $t$, from which we can use \cref{eq:reverse_diff} to compute
$\boldsymbol{x}^{T-1},\dots,\boldsymbol{x}^0$, where~$\boldsymbol{x}^0$ is a newly generated sample conditioned on $\boldsymbol{c}$.

\subsection{Diffusion scene completion}
\label{subsec:diff_scene_comp}

In this work, we use the generative aspect of DDPMs to complete a scene measured in a single scan by a LiDAR sensor.
Similarly to shape completion~\cite{zhou2021iccv-sgac,luo2021cvpr,lyu2022iclr}, the input is a partial point cloud $\mathcal{P} = \{\boldsymbol{p}_1,\dots,\boldsymbol{p}_N\}$
where~${\boldsymbol{p} \in \mathbb{R}^3}$, and the output should be the complete
point cloud ${\mathcal{P}' = \{\boldsymbol{p}'_1,\dots,\boldsymbol{p}'_M\}}$ where ${\boldsymbol{p}' \in \mathbb{R}^3}$. In our case, the partial point cloud is a single LiDAR scan from which we want to achieve scene
completion. Given a sequence of consecutive LiDAR scans and their poses, we can build a map and sample the complete scene ground truth
$\mathcal{G}$ for an individual scan $\mathcal{P}$, where our scene completion $\mathcal{P}'$ should be as close as possible to $\mathcal{G}$.

Given the pair of input scan $\mathcal{P}$ and ground truth $\mathcal{G}$, we can train the DDPM to achieve scene completion.
As detailed in \cref{subsec:ddpm}, we can compute a noisy point cloud $\mathcal{G}^t$ at step $t$ from the complete scene $\mathcal{G}$ in a point-wise fashion:\vspace{-0.04cm}
\begin{equation}
    \label{eq:xt_P}
    \boldsymbol{p}_m^t = \sqrt{\overline{\alpha}_t}\boldsymbol{p}_m + \sqrt{1 - \overline{\alpha}_t}\epsilon,\; \forall \boldsymbol{p}_m \in \mathcal{G},
\end{equation}
\noindent with $\mathcal{G}^t = \{\boldsymbol{p}^t_1,\dots,\boldsymbol{p}^t_M\}$.

In our case, we want to retrieve the complete scene $\mathcal{G}$ from $\mathcal{G}^T$. However, $\mathcal{G}^T$
retains little information from $\mathcal{G}$ due to the $T$ diffusion steps. Therefore, we condition the generation with the scan
$\mathcal{P}$ such that its structure guides the point cloud generation. From \cref{eq:class_free}, the point-wise classifier-free
noise prediction at step $t$ can be rewritten as:
\begin{equation}
    \small
    \label{eq:P_class_free}
    \boldsymbol{\epsilon}'_\theta\hspace{-0.075cm}\left(\mathcal{G}^t,\mathcal{P},t\right) \hspace{-0.05cm} = \hspace{-0.05cm} \boldsymbol{\epsilon}_\theta\hspace{-0.075cm}\left(\mathcal{G}^t,\emptyset,t\right) \hspace{-0.05cm} + \hspace{-0.05cm} s\hspace{-0.075cm}\left[\boldsymbol{\epsilon}_\theta(\mathcal{G}^t,\mathcal{P},t) - \boldsymbol{\epsilon}_\theta\hspace{-0.075cm}\left(\mathcal{G}^t,\emptyset,t\right)\right]\hspace{-0.05cm}.
\end{equation}
For training, at each iteration we select a random step $t \in [0,T]$ and compute $\mathcal{G}^t$ from $\mathcal{G}$ given Gaussian noise ${\boldsymbol{\epsilon} \sim \mathcal{N}\hspace{-0.075cm}\left(\boldsymbol{0},\boldsymbol{I}\right)}$. Then, we use the model to predict
the noise from $\mathcal{G}^t$ conditioned to the LiDAR scan $\mathcal{P}$ or a null token $\emptyset$ given a probability $p$ as in \cref{eq:diff_loss}, supervising with the loss:\vspace{-0.2cm}
\begin{equation}
    \mathcal{L}_{\text{diff}}\hspace{-0.075cm}\left(\mathcal{G}^t,\tilde{\boldsymbol{c}}, t\right) = \big\lVert \boldsymbol{\epsilon} - \boldsymbol{\epsilon}_\theta \hspace{-0.075cm}\left(\mathcal{G}^t, \tilde{\boldsymbol{c}}, t\right) \big\rVert^2,
\end{equation}
\noindent where as in \cref{eq:diff_loss}, ${\tilde{\boldsymbol{c}} \sim \mathcal{B}\hspace{-0.075cm}\left(p\right)}$ with $\mathcal{B}$ as a Bernoulli distribution with outcomes ${\{\emptyset, \mathcal{P}\}}$ with probability $p$ that $\emptyset$ occurs.

During inference, as detailed in \cref{subsec:ddpm}, we can generate a scene conditioned to a LiDAR scan $\mathcal{P}$, by denoising from~$\mathcal{G}^T$ to $\mathcal{G}^0$ which is the predicted completion $\mathcal{P}'$.

\subsection{Local point denoising}
\label{sec:local_point_denoising}

\begin{figure}[t]
  \centering
  \includegraphics[width=0.95\linewidth]{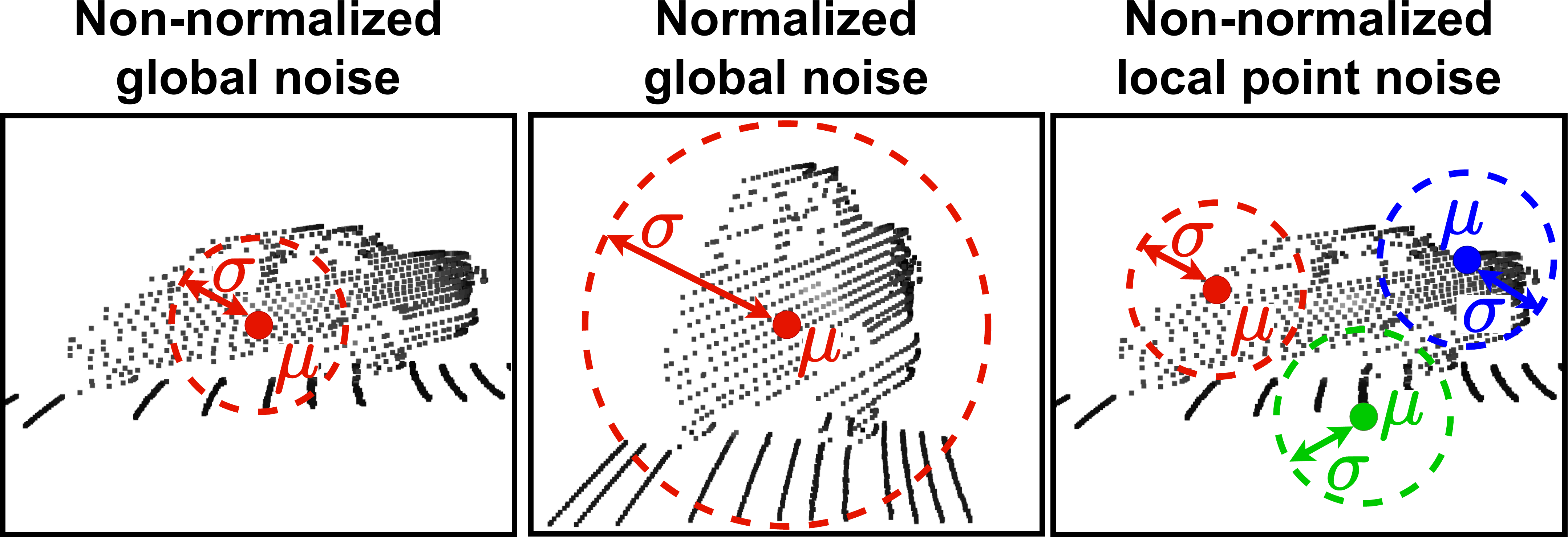}
   \vspace{-0.15cm}
   \caption{Comparison between Gaussian noise with standard deviation $\sigma$ and mean $\mu$ over non-normalized and normalized input point cloud and our proposed local point-wise noise formulation.}
   \label{fig:global_local_noise}
   \vspace{-0.5cm}
\end{figure}

The formulation detailed in \cref{subsec:diff_scene_comp} is usually used for shape completion~\cite{zhou2021iccv-sgac,lyu2022iclr}. Even though achieving promising
results for shape completion, this formulation may not directly work at the scene scale. For single object shapes, the data is either normalized or within a small range close to a Gaussian distribution with
mean ${\mu = \boldsymbol{0}}$ and standard deviation ${\Sigma = \boldsymbol{I}}$. For scene scale, the LiDAR data has a much larger scale, and the data range differs depending on the point
cloud axis. Therefore, the input data distribution is far from a Gaussian distribution $\mathcal{N}(\boldsymbol{0},\boldsymbol{I})$, and if we normalize the data,
we lose many details in the scene due to compressing it into a much smaller range as illustrated in~\cref{fig:global_local_noise}.

To overcome this problem, we reformulate the diffusion process as a point-wise local problem. Instead of sampling~$\boldsymbol{x}^t$ as a mixed distribution between
${\boldsymbol{\epsilon} \sim \mathcal{N}(\boldsymbol{0},\boldsymbol{I})}$ and~$\boldsymbol{x}^0$ as in \cref{eq:closed_form}, we formulate the diffusion process as a noise offset added locally to each point ${\boldsymbol{p}_m \in \mathcal{G}}$. In this case,
from \cref{eq:closed_form}, we set ${\boldsymbol{x}^0 = \boldsymbol{0}}$ and add $\boldsymbol{x}^t$ to $\boldsymbol{p}_m$:
\begin{align}
    \label{eq:local_diff}
    \boldsymbol{p}_m^t&=\boldsymbol{p}_m + \hspace{-0.075cm}\left(\sqrt{\overline{\alpha}_t}\boldsymbol{0}+\sqrt{1 - \overline{\alpha}_t}\boldsymbol{\epsilon}\right), &\\
    \label{eq:local_diff_simpler}
    &= \boldsymbol{p}_m + \sqrt{1 - \overline{\alpha}_t}\boldsymbol{\epsilon}.
\end{align}

With this formulation, the noise $\boldsymbol{\epsilon}$ is a random offset scaled w.r.t. the step $t$ added to each point $\boldsymbol{p}_m$ in $\mathcal{G}$. The model needs to
predict the noise at each step $t$, slowly moving the noisy points towards the target scene $\mathcal{G}$ conditioned to
the LiDAR scan $\mathcal{P}$, still operating in the original scale.

During inference, due to this local diffusion formulation,~$\mathcal{G}^T$ cannot be approximated by a Gaussian distribution. Instead, we can generate $\mathcal{G}^T$ from the
LiDAR scan~$\mathcal{P}$. Besides, to complete the LiDAR scan, we need more
points than the input scan. Therefore, given a single LiDAR scan~$\mathcal{P}$, we increase its size by concatenating its points~$K$
times to get $\mathcal{P}^* = \{\boldsymbol{p}^*_1,\dots,\boldsymbol{p}^*_{KN}\}$, where ${M=KN}$. Then, we sample a Gaussian noise for each point $\boldsymbol{p}^*_m \in \mathcal{P}^*$ and
compute the initial noisy point cloud $\mathcal{P}^T$ from $\mathcal{P}^*$ with \cref{eq:local_diff_simpler}. Finally, we calculate the $T$ denoising steps by predicting
the noise at step $t$ from \cref{eq:class_free}, and denoising it with \cref{eq:reverse_diff} to get the complete scan ${\mathcal{P}' = \mathcal{P}^0}$.

Note that, as long as $\mathcal{P}^T$ is ``noisy enough'' to resemble $\mathcal{G}^T$ as seen during training, the generation process is the same independent of
using $\mathcal{P}^*$ or the ground truth $\mathcal{G}$ to sample the initial $\boldsymbol{x}^T$.

\subsection{Noise prediction regularization}

DDPMs use a leveraged formulation to train the model to predict only the noise added to the data. This formulation
has only to optimize an $\mathcal{L}_2$ loss between the added noise and the model prediction. However, this formulation optimizes the model
to precisely predict the noise added to each point, ignoring the overall distribution of the noise sampled.

\begin{figure}[t]
  \centering
  \includegraphics[width=0.85\linewidth]{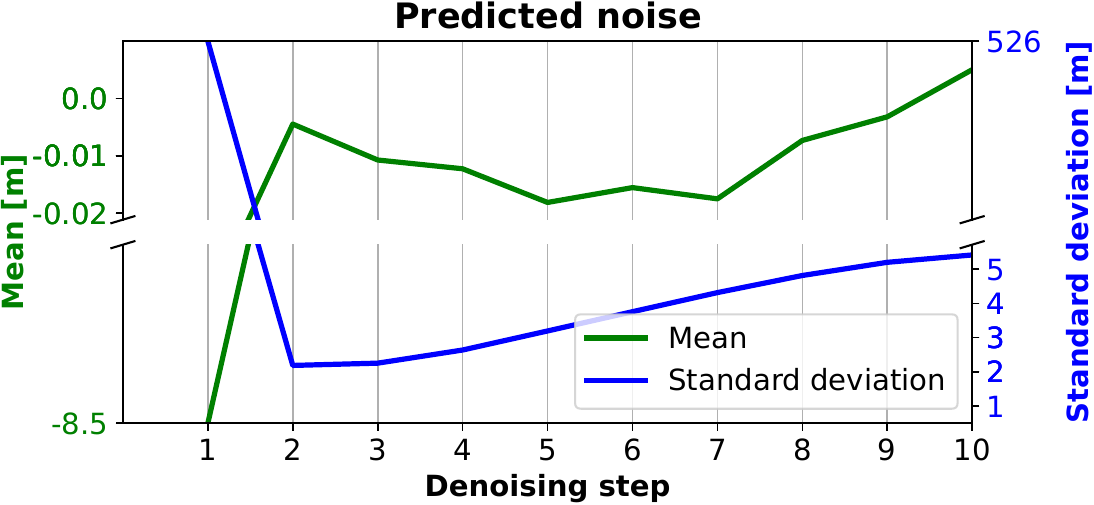}
   \caption{Mean and standard deviation of the predicted noise $\boldsymbol{\epsilon}_\theta$ without the noise regularization. In this experiment we use DPMSolver~\cite{lu2022neurips}
   to reduce the denoising steps from $1,000$ to $10$.}
   \label{fig:pred_noise}
   \vspace{-0.3cm}
\end{figure}

Given that the added noise ${\boldsymbol{\epsilon}\sim\mathcal{N}\hspace{-0.075cm}\left(\boldsymbol{0},\boldsymbol{I}\right)}$,
it is reasonable to expect that the prediction ${\boldsymbol{\epsilon}_\theta\hspace{-0.075cm}\left(\mathcal{G}^t,\mathcal{P},t\right) \approx \mathcal{N}\hspace{-0.075cm}\left(\boldsymbol{0},\boldsymbol{I}\right)}$.
However, the model predicts a peaky distribution far from the expected, as shown in \cref{fig:pred_noise}. The predicted noise
starts with a mean far from zero and with a large standard deviation. As the denoising starts the mean gets closer to zero but the standard
deviation is still far from one. Therefore, we propose a regularization to approximate
$\boldsymbol{\epsilon}_\theta\hspace{-0.075cm}\left(\mathcal{G}^t,\mathcal{P},t\right)$ to~$\mathcal{N}\hspace{-0.075cm}\left(\boldsymbol{0},\boldsymbol{I}\right)$. We compute the mean $\overline{\boldsymbol{\epsilon}}_\theta$
and the standard deviation $\hat{\boldsymbol{\epsilon}}_\theta$ over $\boldsymbol{\epsilon}_\theta\hspace{-0.075cm}\left(\mathcal{G}^t,\mathcal{P},t\right)$ and calculate the regularization
losses:

\begin{equation}
    \mathcal{L}_{\text{mean}} = {\overline{\boldsymbol{\epsilon}}_\theta}^2 \quad \text{and} \quad
    \mathcal{L}_{\text{std}} = \hspace{-0.075cm}\left(\hat{\boldsymbol{\epsilon}}_\theta - 1\right)^2,
\end{equation}

\noindent then our final loss becomes:
\begin{equation}
    \mathcal{L} = \mathcal{L}_{\text{diff}} + r \left(\mathcal{L}_{\text{mean}} + \mathcal{L}_{\text{std}}\right),
\end{equation}

\noindent where $r$ is a weighting factor.

With this regularization, we aim at smoothing the peaky distribution from the predicted noise, and approximating it to the expected distribution,
\ie, ${\boldsymbol{\epsilon}_\theta\hspace{-0.075cm}\left(\mathcal{G}^t,\mathcal{P},t\right) \approx \mathcal{N}\hspace{-0.075cm}\left(\boldsymbol{0},\boldsymbol{I}\right)}$.

\subsection{Refinement network}
\label{sec:refinement_scheme}

Despite the impressive results from DDPMs, the denoising process still demands time since it has to predict all the $T$ steps. Recent
efforts~\cite{salimans2022iclr,meng2023cvpr,song2021iclr,karras2022neurips,lu2022neurips,lu2023arxiv} focus on increasing the inference speed.
However, by reducing the inference time, the generation quality may also decrease. Besides, processing 3D scene-scale data demands many computational resources. This limitation hinders the number
of points we can generate to represent the complete scene. Therefore, we follow the refinement upsampling scheme proposed
by Lyu \etal~\cite{lyu2022iclr}. As done by them, we train an additional model to refine the scene generated by the diffusion process
while upsampling it by predicting $\kappa$ offsets $\boldsymbol{o}_\kappa \in \mathbb{R}^3$ for each point in the completed scene~$\mathcal{P}'$.
Then, we add the offsets to the completed scene points as~${\{\boldsymbol{p}'_n + \boldsymbol{o}_0,\dots,\boldsymbol{p}'_n + \boldsymbol{o}_\kappa\},\; \forall\; \boldsymbol{p}'_n \in \mathcal{P}'}$
refining the diffusion completion while upsampling it.

\subsection{Noise predictor architecture}
\label{sec:noise_pred_descrip}

\begin{figure}[t]
    \centering
    \includegraphics[width=0.9\linewidth]{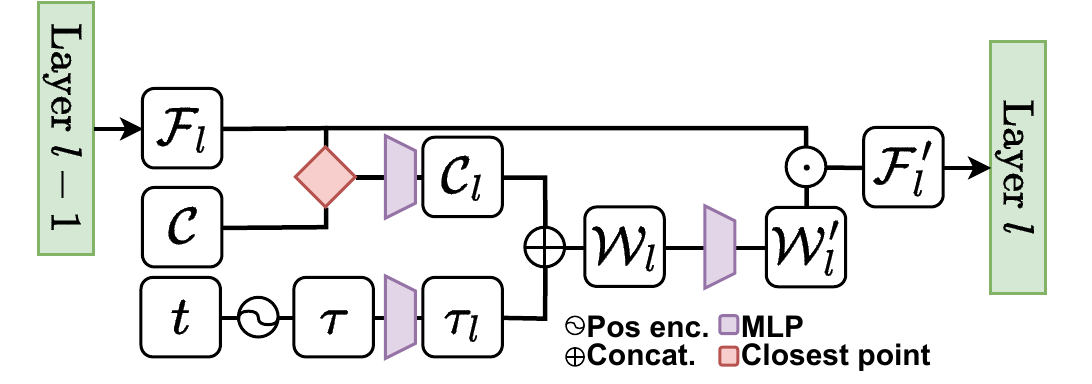}
     \caption{Diagram of the conditioning in each layer $l$.}
     \label{fig:cond_diag}
     \vspace{-0.25cm}
  \end{figure}

We parameterize the denoising process using a MinkUNet~\cite{choy2019cvpr} as the noise predictor which uses sparse convolutions to process 3D data.
To encode information from the conditioning scan $\mathcal{P}$, we use the encoder part from MinkUNet with the same architecture
as the noise predictor. The encoder downsamples $\mathcal{P}$ to a smaller version ${\mathcal{C} = \{\boldsymbol{c}_{n'} \in \mathbb{R}^{d_c} \mid 1 < n' < N'\}}$,
where ${N' < N}$ and $d_c$ is the encoder output embedding size. To encode the denoising step $t$, similar to previous work~\cite{zhou2021iccv-sgac}, we use a
sinusoidal positional encoding to compute the temporal embedding~${\tau \in \mathbb{R}^{d_t}}$. Then, before each layer $l$ in the denoising model, we compute the closest point
between the layer input points ${\mathcal{F}_l = \{\boldsymbol{f}_{n'_l} \in \mathbb{R}^{d_l} \mid 1 < n'_l < N'_l\}}$ and the conditioning embeddings $\mathcal{C}$ to get a per-point guidance,
passing it over an MLP to get ${\mathcal{C}_l = \{\boldsymbol{c}_{n'_l} \in \mathbb{R}^{d_l} \mid 1 < n'_l < N'_l\}}$. Then, we compute ${\tau_l \in \mathbb{R}^{d_l}}$ from $\tau$
through another MLP, and concatenate $\tau_l$ to each point in $\mathcal{C}_l$ to obtain ${\mathcal{W}_l = \{\boldsymbol{w}_{n'_l} \in \mathbb{R}^{2d_l} \mid 1 < n'_l < N'_l\}}$.
Finally, we use one more MLP layer to project $\mathcal{W}_l$ to the layer feature dimension $d_l$ and get $\mathcal{W}'_l$.
Then, we compute ${\mathcal{F}'_l = \mathcal{W}'_l \odot \mathcal{F}_l}$ as an element-wise multiplication, which is then feed as the input to layer $l$, as depicted in \cref{fig:cond_diag}. As the refinement network, we use the same MinkUNet architecture used for the noise predictor without the conditioning encoder.
For more details on the embeddings dimensions, noise predictor and refinement network architectures, we refer to the supplementary material.

\section{Experiments}
\label{sec:experiments}

\textbf{Datasets.} For training our DDPM, we used the SemanticKITTI dataset~\cite{behley2019iccv,geiger2012cvpr}, an autonomous driving
benchmark with point-wise annotations over sequences of LiDAR scans collected in an urban environment. To generate the ground truth complete scans,
we used the dataset poses to aggregate the scans in the sequence and remove moving objects with the semantic labels, building a map for each sequence. For evaluation, we used the validation set
from SemanticKITTI, \ie, sequence $08$. Additionally, we used sequence $00$ from the KITTI-360 dataset~\cite{liao2022pami} and collected our own data
with an Ouster LiDAR OS-1 with 128 beams to further compare the approaches.

For SemanticKITTI and KITTI-360, we used the ground truth poses to build the map, and for our data,
we used KISS-ICP~\cite{vizzo2023ral} to get the scan poses for our sequence. To remove the moving objects from the map in KITTI-360 and our data, we
used an off-the-shelf moving object segmentation~\cite{mersch2023ral}. To compute the evaluation metrics, for each scan in the sequences, we remove the
moving objects using the semantic labels using only the static points as input to the scene completion methods. Then, we evaluate the completed scene
by comparing it with the corresponding region in the ground truth map.

\textbf{Training.} We train our model for $20$ epochs, using~only the training set from SemanticKITTI. As optimizer, we used Adam~\cite{kingma2015iclr} with a
learning rate of $10^{-4}$ decreased by half every $5$ epochs, and decay of~$10^{-4}$, with batch size equal to $2$. For the diffusion parameters, we used ${\beta_0=3.5 \cdot 10^{-5}}$ and ${\beta_T=0.007}$,
with the number of diffusion steps ${T = 1000}$, linearly interpolating between $\beta_0$ and $\beta_T$ to define $\beta_1,\dots,\beta_{T-1}$.
We set the noise regularization ${r = 5.0}$, and the classifier-free probability ${p = 0.1}$. For the MinkUNet parameters, we set
the quantization resolution to $0.05\,$m. For each input scan, we define the scan range as $50\,$m and sample $18,000$ points with farthest point sampling. For the ground
truth, we randomly sample $180,000$ points without replacement. For the refinement network we use ${\kappa=6}$ as the number of offsets.

\textbf{Inference.} During inference, we use DPMSolver proposed by Lu \etal~\cite{lu2022neurips}, reducing the number of denoising steps $T$ from $1,000$ to $50$.
Besides, we set the classifier-free conditioning weight to $s = 6.0$. To maintain the same amount of points used during training, we again use the scan
max range as $50\,$m and sample $18,000$ points with farthest point sampling. Furthermore, as explained in \cref{sec:local_point_denoising}, we set
$K=10$ to define the input noisy scan $\mathcal{P}^*$.

\textbf{Baselines.} We compare our method with different scene completion methods, LMSCNet~\cite{roldao2020threedv}, PVD~\cite{zhou2021iccv-sgac}, Make It Dense (MID)~\cite{vizzo2022ral}, and LODE~\cite{li2023icra-llce}.
For all baselines, we used their official code and the provided weights also trained on SemanticKITTI. For PVD, we trained the approach with SemanticKITTI with their
default parameters. We also follow their data loading, where the point clouds are normalized before the diffusion process. LMSCNet~\cite{roldao2020threedv} and LODE~\cite{li2023icra-llce} are limited
to a fixed voxel grid of $51.2\,\text{m}\times51.2\,\text{m}\times6.4\,\text{m}$. Given that our point cloud generation is done over a scan with a radius of $50\,$m, we divide
the input scan into four quadrants over the $360^{\circ}$ LiDAR field of view, generating the complete scene over each quadrant and finally gathering them together as the
final prediction. All baselines and our method were trained only with SemanticKITTI, and later evaluated on SemanticKITTI, and on KITTI-360 and our data without fine-tuning.

\subsection{Scene reconstruction}
\label{sec:scene_reconstruction}

In this experiment we evaluate how close is the predicted scan completion from the expected complete scene. To do so, we quantify it with two metrics,
the Chamfer distance (CD) and the Jensen-Shannon divergence (JSD). The Chamfer distance evaluates the completion at point level, measuring
the level of detail of the generated scene by calculating how far are its points from the expected scene. The JSD compares the points distribution between the generated and the ground truth scene.
For the JSD, we follow the evaluation done by Xiong \etal~\cite{xiong2023cvpr-lcrf}, where the scene is first voxelized with a grid resolution of $0.5\,$m and then projected to a bird's eye view (BEV)
evaluating over this projection.

\begin{table}
  \centering
  \small
  \begin{tabular}{lcc}
    \toprule 
    Method & CD [m] $\downarrow$ & $\text{JSD}_{\text{BEV}}$ [m] $\downarrow$ \\ 
    \midrule 
    LMSCNet~\cite{roldao2020threedv} & 0.641 & 0.431 \\ 
    LODE~\cite{li2023icra-llce} & 1.029 & 0.451 \\ 
    MID~\cite{vizzo2022ral} & 0.503 & 0.470 \\ 
    PVD~\cite{zhou2021iccv-sgac} & 1.256 & 0.498 \\ 
    \midrule
    Ours & 0.434 & 0.444 \\ 
    Ours refined & \textbf{0.375} & \textbf{0.416} \\ 
    \bottomrule
  \end{tabular}
  \caption{Mean chamfer distance and Jensen-Shannon divergence evaluation on validation set from SemanticKITTI.}
  \label{tab:chamfer_dist_jsd}
  \vspace{-0.3cm}
\end{table}

\begin{table}
  \centering
  \resizebox{1.0\linewidth}{!}{
  \setlength{\tabcolsep}{0.1cm}
  \begin{tabular}{lcc|cc}
    \toprule 
    & \multicolumn{2}{c|}{KITTI-360} & \multicolumn{2}{c}{Our data} \\
    \cmidrule(lr){2-3}\cmidrule(lr){4-5}
    Method & CD [m] $\downarrow$ & $\text{JSD}_{\text{BEV}}$ [m] $\downarrow$ & CD [m] $\downarrow$ & $\text{JSD}_{\text{BEV}}$ [m] $\downarrow$ \\ 
    \midrule 
    LMSCNet~\cite{roldao2020threedv} & 0.979 & 0.496  & 0.826 & 0.439 \\ 
    LODE~\cite{li2023icra-llce} & 1.565 & 0.483  & \textbf{0.387} & 0.389 \\ 
    MID~\cite{vizzo2022ral} & 0.637 & 0.476  & 0.475 & 0.379 \\ 
    \midrule
    Ours & 0.564 & 0.459  & 0.518 & 0.360 \\ 
    Ours refined & \textbf{0.517} & \textbf{0.446}  & 0.471 & \textbf{0.341} \\ 
    \bottomrule
  \end{tabular}
  }
  \caption{Mean chamfer distance and Jensen-Shannon divergence evaluation on KITTI-360 sequence $00$ and our data.}
  \label{tab:chamfer_dist_jsd_more}
  \vspace{-0.43cm}
\end{table}

\begin{figure*}[t]
  \centering
  \includegraphics[width=1.0\linewidth]{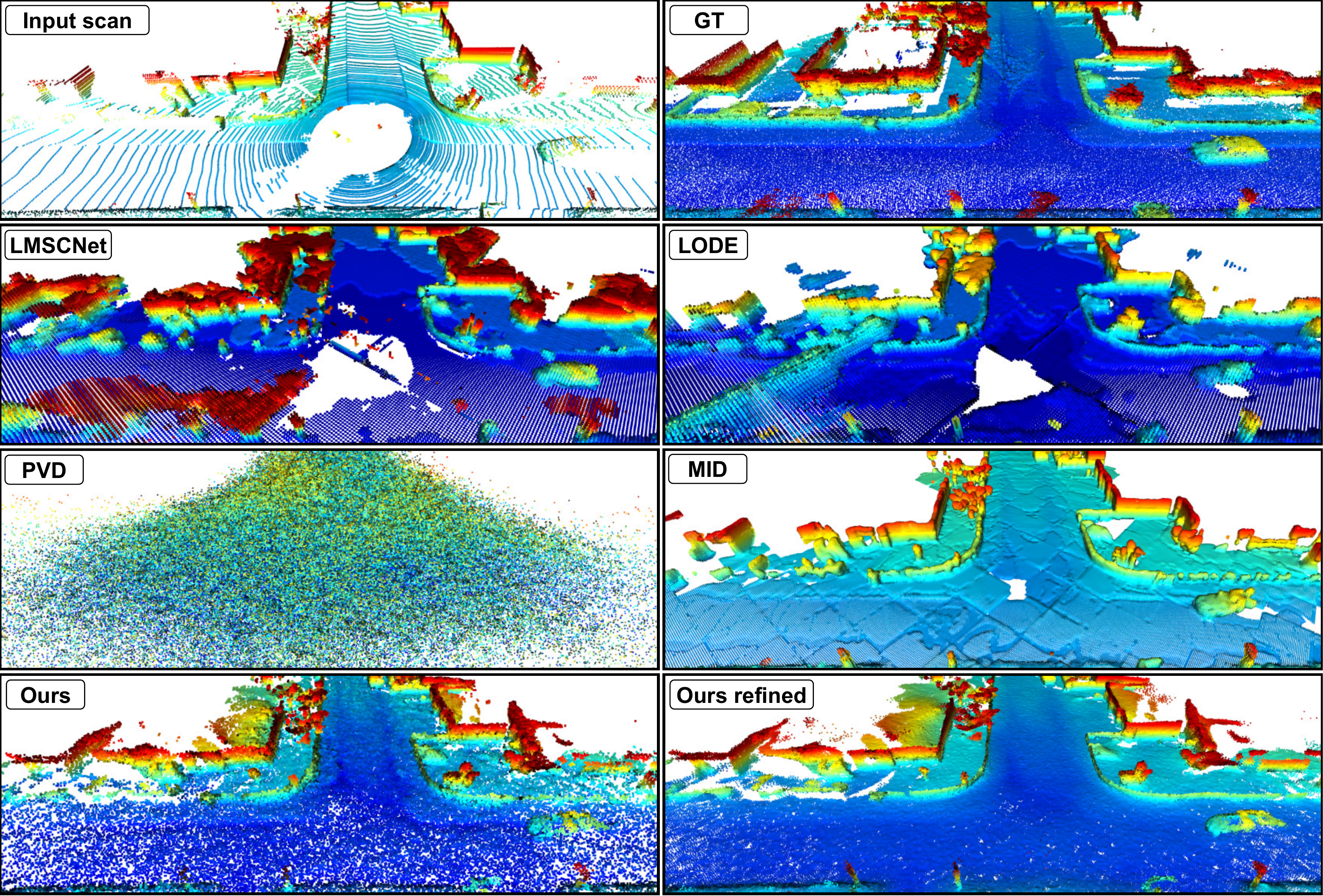}
   \vspace{-0.6cm}
   \caption{Qualitative results on one scan from KITTI-360. Colors depict point height normalized by the height range of each point cloud.}
   \label{fig:quali_360}
   \vspace{-0.3cm}
\end{figure*}

\cref{tab:chamfer_dist_jsd} shows the results comparing our approach with previous state-of-the-art methods, where our method achieves the best performance in both metrics.
First, we can notice that the state-of-the-art shape generation diffusion method, PVD, achieves the lowest performance, showing that current 3D diffusion methods cannot directly be applied to
scene-scale data. The best performance of our method over the CD metric can be explained by the fact that our method operates directly on the points, which enables it to produce a more detailed scene compared
to the baselines. The scene representation from SDF-based methods inherits artifacts from the surface approximation and voxelization, impacting the details in the reconstructed scene and therefore decreasing
their performance with respect to the CD. The JSD evaluates the reconstructed scene points distribution over a voxelized grid
comparing the overall scene distribution between the generated and the expected completion. Even though our method is not optimized over a voxel representation, we still achieve the best performance,
showing that our scene completion is at the same time closer to the expected point distribution and can yield more details.

\cref{tab:chamfer_dist_jsd_more} compares the results of the scene completion methods on KITTI-360 and our collected data. Due to the 
poor performance of PVD over KITTI dataset, we do not evaluate it on those datasets. For KITTI-360 we notice
the same behavior as in \cref{tab:chamfer_dist_jsd}, where our method achieves the best performance in both metrics.
When evaluating in our data, the performance of the SDF-based methods improve. This is expected since our data has denser point clouds, which
is an advantage for such methods since they rely on the input points to approximate a surface to represent the scene. However, our method still
achieves the best performance on the JSD metric and competitive performance on the CD metric. This evaluation shows that our method
can still achieve scene completion over different datasets without fine-tuning since its generation is conditioned to the input scan. In \cref{fig:quali_360}
we can compare the scene completion generation between the methods. We can see that the diffusion baseline, PVD, fails on generating scene-scale data.
SDF-based methods inherits artifacts from the voxelization, while our method, especially after the refinement, can generate a scene closer to the expected, following closely the
structural information from the input scan.

\subsection{Scene occupancy}
\label{sec:voxel_eval}

\begin{table}
  \centering
  \small
  \begin{tabular}{lccc}
    \toprule 
    & \multicolumn{3}{c}{IoU [\%]} \\ 
    & \multicolumn{3}{c}{Grid resolution (m)} \\ 
    \cmidrule(lr){2-4}
    Method & 0.5 & 0.2 & 0.1 \\ 
    \midrule 
    LMSCNet~\cite{roldao2020threedv} & 32.23 & 23.05 & 3.48 \\ 
    LODE~\cite{li2023icra-llce} & 43.56 & \textbf{47.88} & 6.06 \\ 
    MID~\cite{vizzo2022ral} & \textbf{45.02} & 41.01 & 16.98 \\ 
    PVD~\cite{zhou2021iccv-sgac} & 21.20 & 7.96 & 1.44 \\ 
    \midrule
    Ours & 42.49 & 33.12 & 11.02 \\ 
    Ours refined & 40.71 & 38.92 & \textbf{24.75} \\ 
  \bottomrule
  \end{tabular}
  \caption{Completion metric where the IoU is computed against the ground truth and prediction grids with different resolutions.}
  \label{tab:completion_diff_res}
  \vspace{-0.4cm}
\end{table}

In this experiment, we assess the scene completion by evaluating the occupancy of the predicted scene compared with the ground truth. To do so, we follow the evaluation proposed by
Song \etal~\cite{song2017cvpr} where the intersection-over-union (IoU) is computed between the predicted and ground
truth voxelized scene, classifying each voxel as occupied or not. In this evaluation, we compute the IoU at three different voxel resolutions,
\ie, $0.5\,$m, $0.2\,$m, and $0.1\,$m. With a voxel size of $0.5\,$m, we evaluate the occupancy over the coarse scene, where the scene details are not considered. As we decrease the voxel size,
more fine-grained details are considered in the evaluation.

\begin{table}
  \centering
  \resizebox{1.0\linewidth}{!}{
  \begin{tabular}{lccc|ccc}
    \toprule 
    & \multicolumn{3}{c|}{KITTI-360 (IoU) [\%]} & \multicolumn{3}{c}{Our data (IoU) [\%]}  \\
    & \multicolumn{3}{c|}{Grid resolution (m)} & \multicolumn{3}{c}{Grid resolution (m)} \\ 
    \cmidrule(lr){2-4}\cmidrule(lr){5-7}
    Method & 0.5 & 0.2 & 0.1 & 0.5 & 0.2 & 0.1 \\ 
    \midrule 
    LMSCNet~\cite{roldao2020threedv} & 25.46 & 16.35 & 2.99  & 21.93 & 8.48 & 0.95 \\ 
    LODE~\cite{li2023icra-llce} & 42.08 & \textbf{42.63} & 5.85  & 42.99 & 42.24 & 5.14 \\ 
    MID~\cite{vizzo2022ral} & \textbf{44.11} & 36.38 & 15.84  & \textbf{44.47} & \textbf{44.08} & 16.38 \\ 
    \midrule
    Ours & 42.22 & 32.25 & 10.80  & 37.16 & 29.17 & 6.53 \\ 
    Ours refined & 40.82 & 36.08 & \textbf{21.34}  & 38.51 & 40.20 & \textbf{17.48} \\ 
    \bottomrule
    
  \end{tabular}
  }
  \caption{Completion metric where the IoU is computed against the ground truth and prediction grids with different resolutions.}
  \label{tab:completion_diff_res_more}
  \vspace{-0.55cm}
\end{table}

\cref{tab:completion_diff_res} shows the IoU of our method compared to the baselines at the different voxel resolutions. 
First, the diffusion baseline PVD has the lowest performance overall. This again shows that current state-of-the-art 3D shape completion diffusion
methods cannot be directly applied to scene-scale data. At a higher voxel size, our approach stays behind some SDF-based baselines. This is reasonable
in this evaluation since SDF methods use a voxel representation to reconstruct the scene. Therefore, its reconstruction is equally
distributed over the point cloud and its voxel representation is denser compared to our result voxelized. As we decrease the voxel size, the baselines
performance drops. At the lowest resolution, our method outperforms the baselines. LMSCNet~\cite{roldao2020threedv} and LODE~\cite{li2023icra-llce} are
bound to a voxel resolution of $0.2\,$m, therefore with a voxel size of $0.1\,$m their performance drops drastically.
Make It Dense~\cite{vizzo2022ral} was trained with a voxel size of $0.1\,$m, however, our method still outperforms it at this resolution.
This shows the advantage of our approach. Since it is trained at point level, it can produce a more detailed scene, not limited
to a fixed grid~size.

Due to the poor performance of PVD on SemanticKITTI, we compared our method only with non-diffusion approaches
for the other two datasets. In \cref{tab:completion_diff_res_more}, the same behavior is seen on KITTI-360 and our data. At higher voxel resolution, the
SDF baselines have a higher IoU, while with a lower voxel size, our method achieves the best performance. It is also noteworthy that despite of SDF-based
having advantage in our data as discussed in \cref{sec:scene_reconstruction}, our method still achieves the best performance at lower resolution.
This suggests that our approach can reconstruct the scene with more details, and it is able to generate data from a different dataset than the one it was
trained on, since its generation is guided by the input LiDAR scan.

\subsection{Noise regularization}

\begin{figure}[t]
  \centering
  \includegraphics[width=0.9\linewidth]{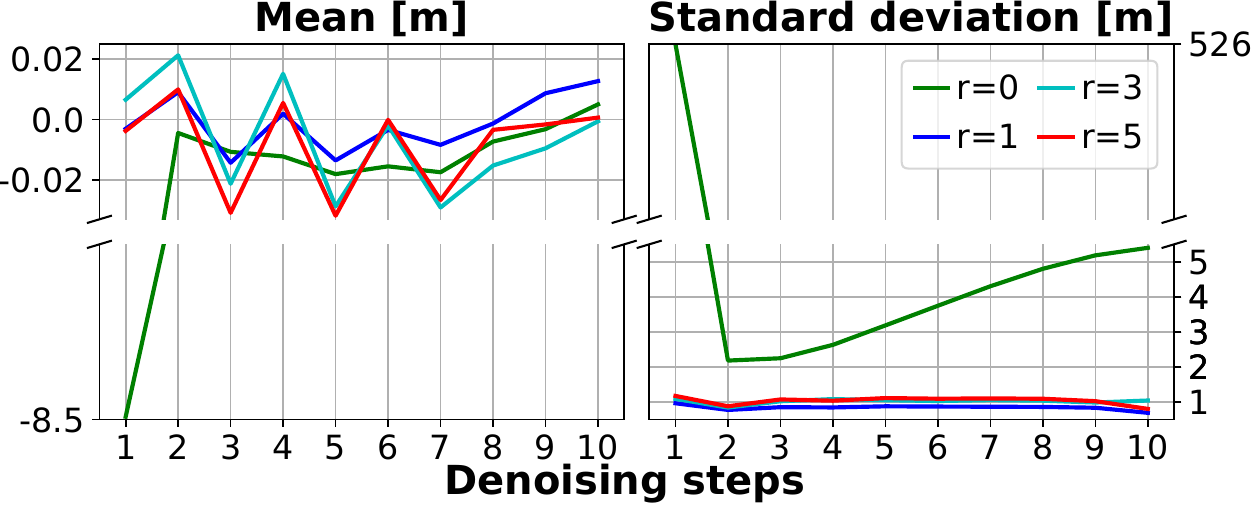}
   \vspace{-0.1cm}
   \caption{Mean and standard deviation of the predicted noise $\boldsymbol{\epsilon}_\theta$ over different regularization weights. In this experiment we use DPMSolver~\cite{lu2022neurips}
   to reduce the denoising steps from $1,000$ to $10$.}
   \label{fig:reg_weights}
   \vspace{-0.5cm}
\end{figure}

In this section, we evaluate the impact of the proposed noise prediction regularization
on the generated scene. We compare the predicted noise distribution with different regularization weights $r$ in \cref{fig:reg_weights} from the $10$ denoising
steps in one scan as in \cref{fig:pred_noise}. As can be seen, without the regularization, \ie, $r = 0$, the predicted noise starts far from the expected distribution, with a mean of around
$-9.0$ and a standard deviation of about $526$. As we denoise the input, the distribution gets closer to the expected, however, still with a high standard deviation. When we add our
proposed regularization, the model already starts predicting a more reasonable noise distribution from the beginning, stabilizing the denoising process. From
this evaluation, we noticed that using $r = 5.0$ achieved a more stable distribution over the denoising steps. In our supplementary material, we provide also
qualitative comparison between the generated point clouds with different regularization weights.

\begin{table}
  \centering
  \small
  \begin{tabular}{lcccc}
    \toprule 
    $r$ & 0.0 & 1.0 & 3.0 & 5.0 \\
    \midrule
    CD [m] & 0.529 & 0.470 & 0.441 & 0.445 \\
    \bottomrule
  \end{tabular}
  \vspace{-0.15cm}
  \caption{Mean chamfer distance over a short sequence from the validation set of SemanticKITTI.}
  \label{tab:chamfer_dist_reg_weights}
  \vspace{-0.5cm}
\end{table}

To evaluate how the regularization impacts the data generation, we compare the model performance over a short sequence from the SemanticKITTI validation set.
We run the scene completion pipeline every one hundred scans without using the refinement network, evaluating only the regularization influence over the noise predictor.
In \cref{tab:chamfer_dist_reg_weights}, we compute the chamfer distance to compare the impact of the regularization over the quality of the generated scene.
As we increase the regularization, the generation quality improves. Despite $r = 3.0$ achieving a slightly better result in this evaluation, we stick with
$r = 5.0$ due to the analysis of the noise distribution from \cref{fig:reg_weights}, and from the qualitative comparisons provided in the supplementary
material.

\vspace{-0.1cm}
\section{Conclusion}

In this paper, we propose a novel point-level denoising diffusion probabilistic model to achieve scene completion using autonomous driving data.
We exploit the generative capabilities of DDPMs to generate the missing parts from a single sparse LiDAR scan. We reformulate the diffusion process
as a local problem. We define each point as the origin of the sampled Gaussian noise, learning an iterative denoising process to gradually
predict offsets to reconstruct the scene from the input noisy LiDAR scan. This formulation enables the processing of scene-scale 3D data,
retaining more details during the denoising process. In our experiments, we compare our method with recent state-of-the-art diffusion and non-diffusion
methods. Our results show that our approach produces a more fine-grained completion compared to the baselines and can achieve scene completion on different
datasets since its generation is conditioned to the input LiDAR scan. Besides, our proposed diffusion formulation
distinguishes from previous state-of-the-art diffusion approaches by enabling the generation of scene-scale 3D data. Furthermore, we
believe that our scene-scale diffusion formulation can support further research in the 3D diffusion generation research field. 

\textbf{Limitations.} Even though achieving compelling results on scene completion, our method is still not able to generate unconditional data. This limits
the data generation capability since it requires an input scan to guide the generation.
In our supplementary material, we show examples of the unconditional generation of our approach. For future work, we plan on extending our method to
generate unconditional data, creating novel 3D point cloud scenes.

\small \textbf{Acknowledgments}. This work has partially been funded by the Deutsche Forschungsgemeinschaft (DFG, German Research Foundation) under Germany’s Excellence Strategy, EXC-2070 – 390732324 – PhenoRob.

{
    \small
    \bibliographystyle{ieeenat_fullname}
    \bibliography{glorified,new}
}

\end{document}


\appendix
\maketitlesupplementary

This supplementary material provides further detailed information on our proposed approach. We provide detailed information on the used network
architectures, and ablations over the different hyperparameters of the generation process, \ie, conditioning weight $s$ and regularization
weight $r$. \cref{sec:diagrams} provides detailed information about the noise predictor and refinement network architectures
and more detailed information about the refinement network training. \cref{sec:reg_abl} gives further ablations over the noise predictor
regularization weight $r$. \cref{sec:cond_abl} presents qualitative and quantitative comparisons
between the scene completion with different conditioning weights~$s$ and the unconditional generation, \ie, $s=0.0$. \cref{sec:denoising_steps} compares
qualitatively the scene completion with different number of denoising steps. Finally, \cref{sec:qual_res} shows further qualitative results
comparing our scene completion with the evaluated baselines. Furthermore, we provide our code within this supplementary material, which we
will make publicly available upon acceptance of the paper.

\begin{figure*}[t]
    \centering
    \vspace{1.5cm}
    \includegraphics[width=1.0\linewidth]{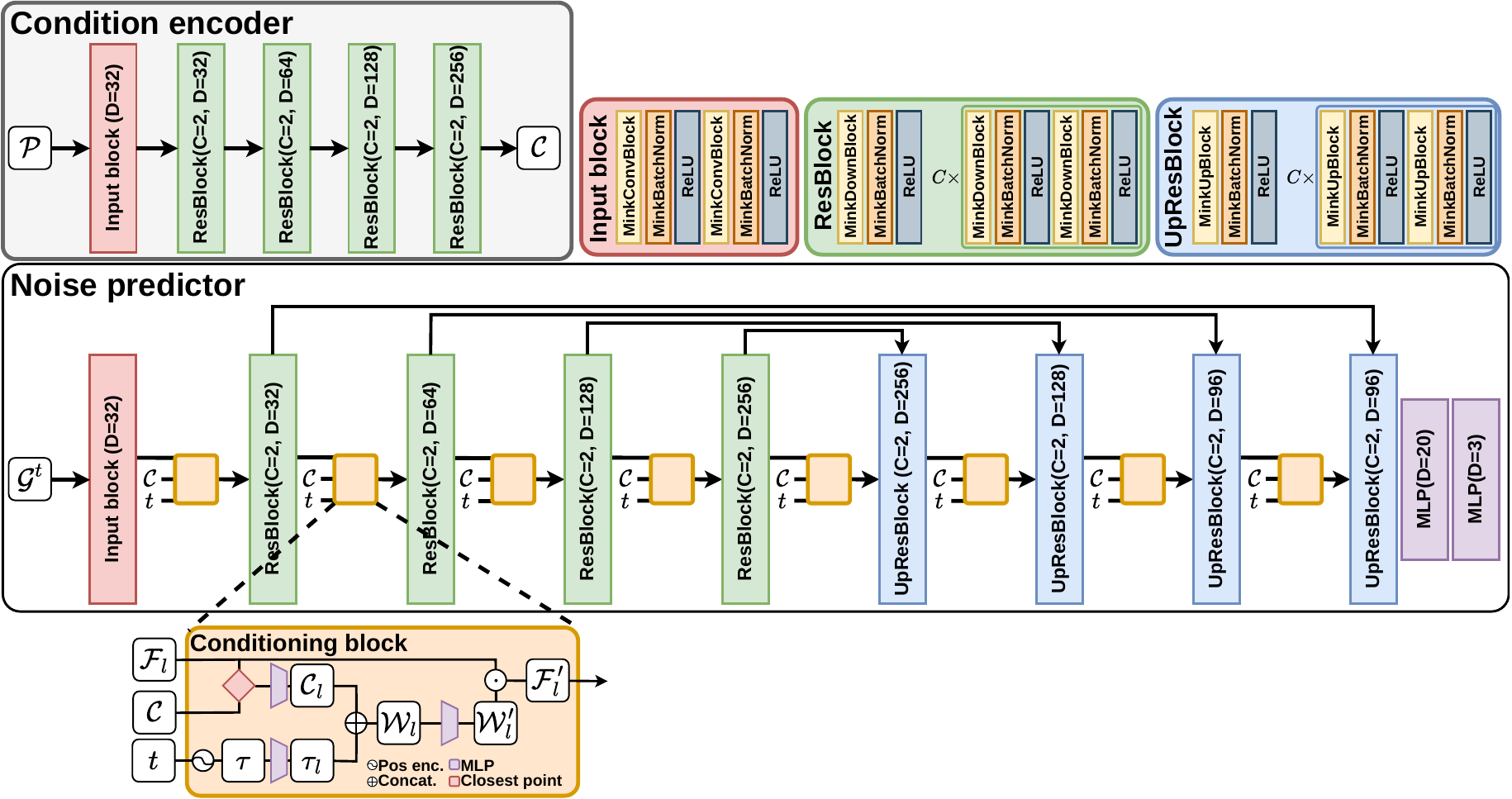}
     \caption{Noise predictor and condition encoder models architecture. The condition encoder receives the scan $\mathcal{P}$ and computes the
     conditioning point cloud $\mathcal{C}$. From $t$, we compute the positional embedding $\tau$ with a dimension $d_t=96$. At each layer $l$, we give
     $\mathcal{C}$ and $\tau$ to the conditioning block together with the layer input features $\mathcal{F}_l$ to get $\mathcal{F}'_l$,
     which is then feed as input to the layer $l$.}
     \label{fig:noise_predictor}
     \vspace{1.2cm}
\end{figure*}

\begin{figure*}[t]
    \centering
    \includegraphics[width=0.65\linewidth]{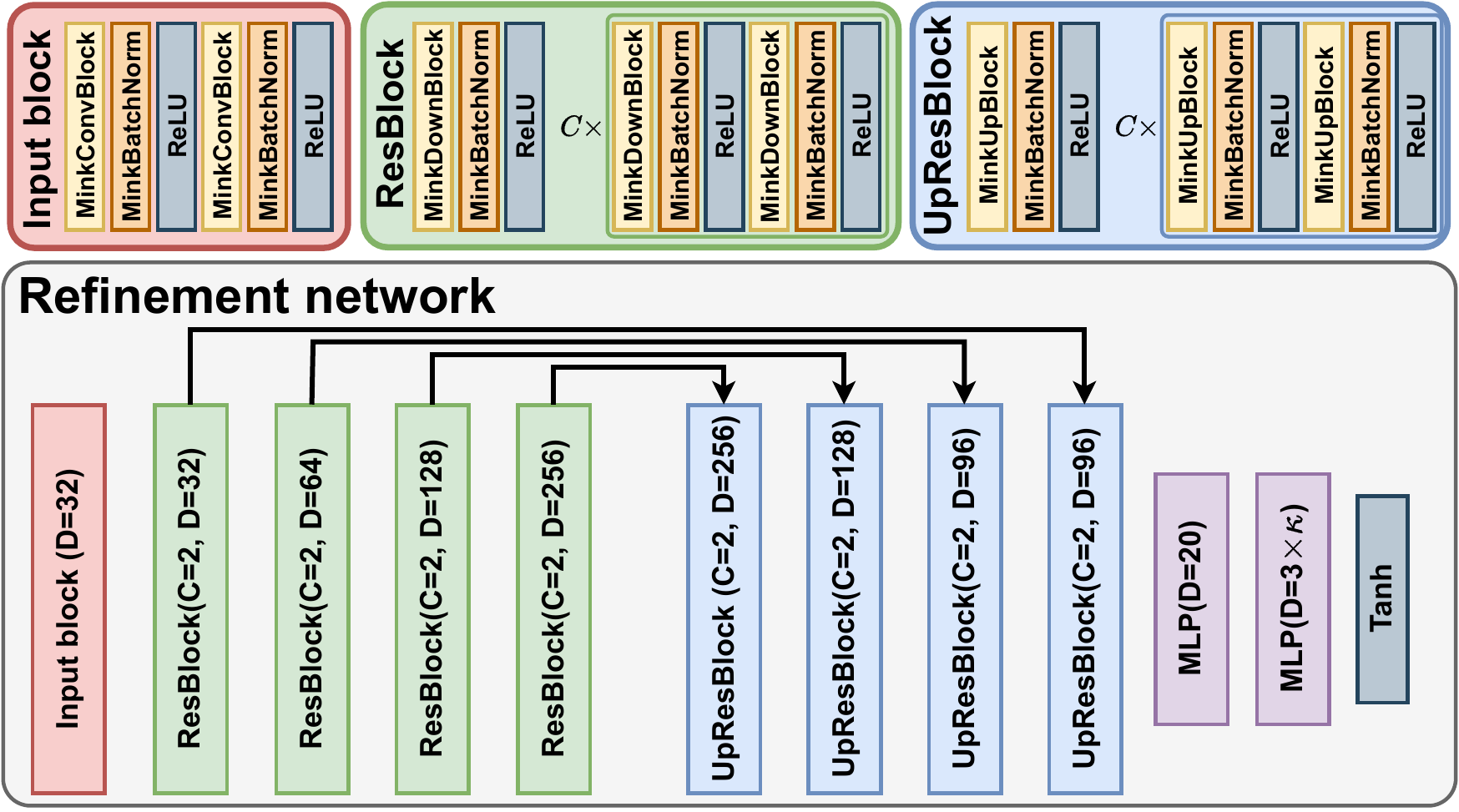}
     \caption{Refinement network architecture.}
     \label{fig:refine_net}
\end{figure*}

\section{Architectures}
\label{sec:diagrams}

This section shows the model architectures for the noise predictor and the refinement network with further details on the training procedure.
\cref{sec:noise_pred} shows the diagram of the noise predictor model together with the condition encoder and how the noise prediction is conditioned
to it. \cref{sec:refine_upsample} presents the refinement upsample network architecture and provides further details on the refinement network
training.

\subsection{Noise predictor}
\label{sec:noise_pred}

As the noise predictor, we used a MinkUNet~\cite{choy2019cvpr} to predict the noise over each point. For the condition encoder, we  used only the
encoder part of the MinkUNet with the same architecture as the noise predictor. As described in \mainref{3.6} of the main paper, before each layer $l$, we compute the positional embeddings
$\tau$ from the denoising step $t$ with an embedding dimension $d_t = 96$, conditioning the layer input $\mathcal{F}_l$ to $\mathcal{C}$ and
$t$ with the conditioning block. \cref{fig:noise_predictor} depicts the noise predictor and condition encoder architecture, with each layer $l$ features dimension
$d_l$ and the conditioning scheme.

\subsection{Refinement upsample network}
\label{sec:refine_upsample}

As the refinement network, we have used the same architecture as the noise predictor with a $\tanh$ activation as the final layer, as depicted in \cref{fig:refine_net}. Given that the
refinement network has to predict just an offset around the diffusion generation, we use a $\tanh$ layer to limit the offset size, avoiding
the model predicting too large offsets.

As mentioned in \mainref{3.5} of the main paper, we used the refinement and upsample scheme proposed by Lyu \etal~\cite{lyu2022iclr}. We train the refinement
model using Adam~\cite{kingma2015iclr} optimizer, with a learning rate of $10^{-4}$ and decay of~$10^{-4}$, with a batch size equal to $8$, training for $5$ epochs.
To generate the refinement ground truth, we aggregate $20$ scans before and $20$ scans after each scan in the training set, using the relative poses between the scans.
We use these aggregated scans as the ground truth $\mathcal{O}_\text{gt}$, and as the input, we copy $\mathcal{O}_\text{gt}$ and add random point
jittering to each point, defining the input $\mathcal{O}$. Then, the model is trained to predict $3\times \kappa$ values for each point, corresponding to
$\kappa$ offsets. We add the $\kappa$ offsets to each point in $\mathcal{O}$, getting the
upsampled refined prediction $\mathcal{O}'$, and supervise it with the symmetric chamfer distance loss $\mathcal{L}_\text{refine}$ as:
\begin{align}
    \mathcal{L}_\text{CD} \left(\mathcal{A}, \mathcal{B}\right) &= \frac{1}{\mid \mathcal{A} \mid} \sum_{\boldsymbol{a} \in \mathcal{A}} \min_{\boldsymbol{b} \in \mathcal{B}} \lVert \boldsymbol{a} - \boldsymbol{b}  \rVert^2_2, \label{eq:cd_refine}\\
    \mathcal{L}_{\text{refine}} &= \mathcal{L}_\text{CD}(\mathcal{O}_\text{gt}, \mathcal{O}') + \mathcal{L}_\text{CD}(\mathcal{O}', \mathcal{O}_\text{gt}).
    \label{eq:cd_refine_loss}
\end{align}

With \cref{eq:cd_refine_loss}, we train the refinement model to predict~$\kappa$ offsets to the input $\mathcal{O}$ such that the upsampled refined
prediction $\mathcal{O}'$ gets as close as possible to the ground truth. With this refinement model, we can generate the scene completion with
our diffusion model with fewer denoising steps using the DPMSolver~\cite{lu2022neurips} and refine it. As mentioned in \mainref{3.5} of the main paper,
with fewer denoising steps, the generation quality may decrease. Therefore, with this refinement network, we can compensate for this lower generation
quality while also upsampling our generated scene completion.

\section{Regularization ablation}
\label{sec:reg_abl}

\begin{figure*}[t]
    \centering
    \includegraphics[width=1.0\linewidth]{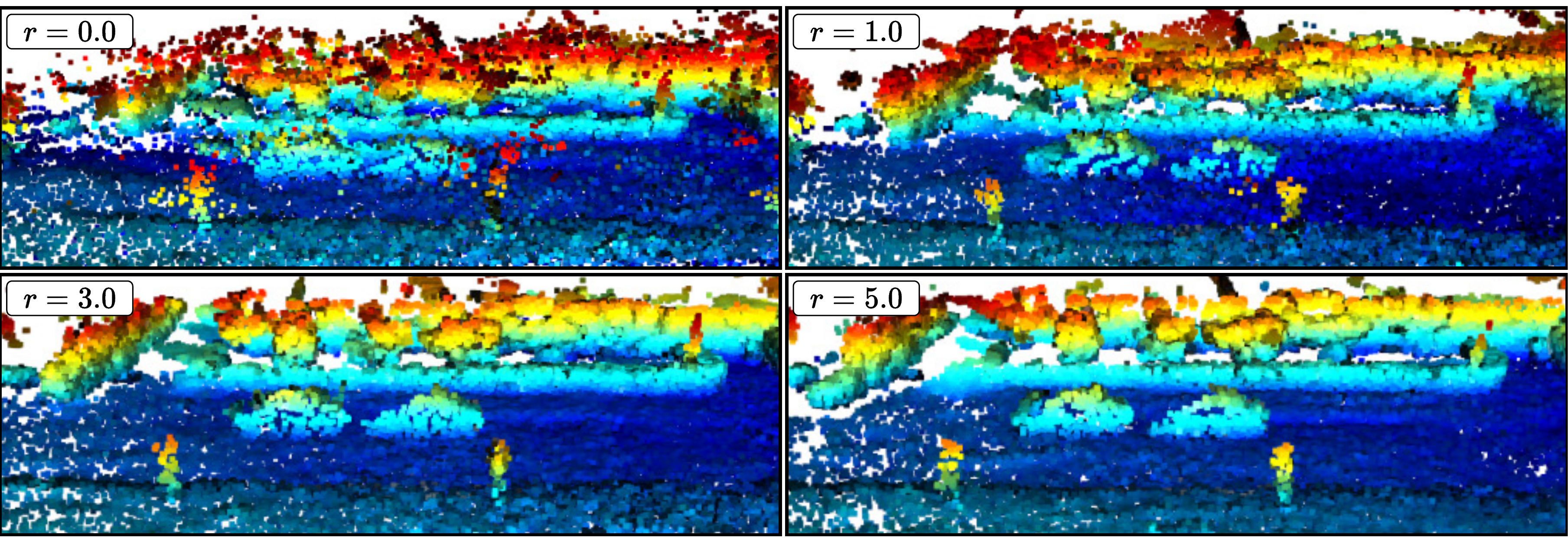}
     \caption{Comparison between results with different regularization weights $r$.}
     \label{fig:reg_weights}
\end{figure*}

This section compares the results of the noise predictor trained with different regularization weights $r$. \cref{fig:reg_weights} compares the scene completion
with the noise predictor trained with different regularization weights. With $r=0.0$, the model can generate structural information with a noisy aspect,
and, in this example, the points from the two parked cars are mixed together without a clear boundary. With $r=1.0$, a less noisy scene completion is generated, but
still, the surfaces in the structure present a noisy aspect. When comparing $r=3.0$ and $r=5.0$, both generated scene depicts a more detailed and
less noisy scene, compared with lower regularization weights $r$. However, using $r=5.0$ achieves more fine-grained structural details. The surfaces in the scene
appear to have a flatter aspect, and the sidewalk curbs seem better defined. Also, the two parked cars retain more details, \eg, the windows space. Given
this analysis and the quantitative results present in Tab. 5 of the main paper, we use $r=5.0$ in the main experiments.

\section{Condition weights ablation}
\label{sec:cond_abl}
\begin{figure*}[p]
    \centering
    \includegraphics[width=1.0\linewidth]{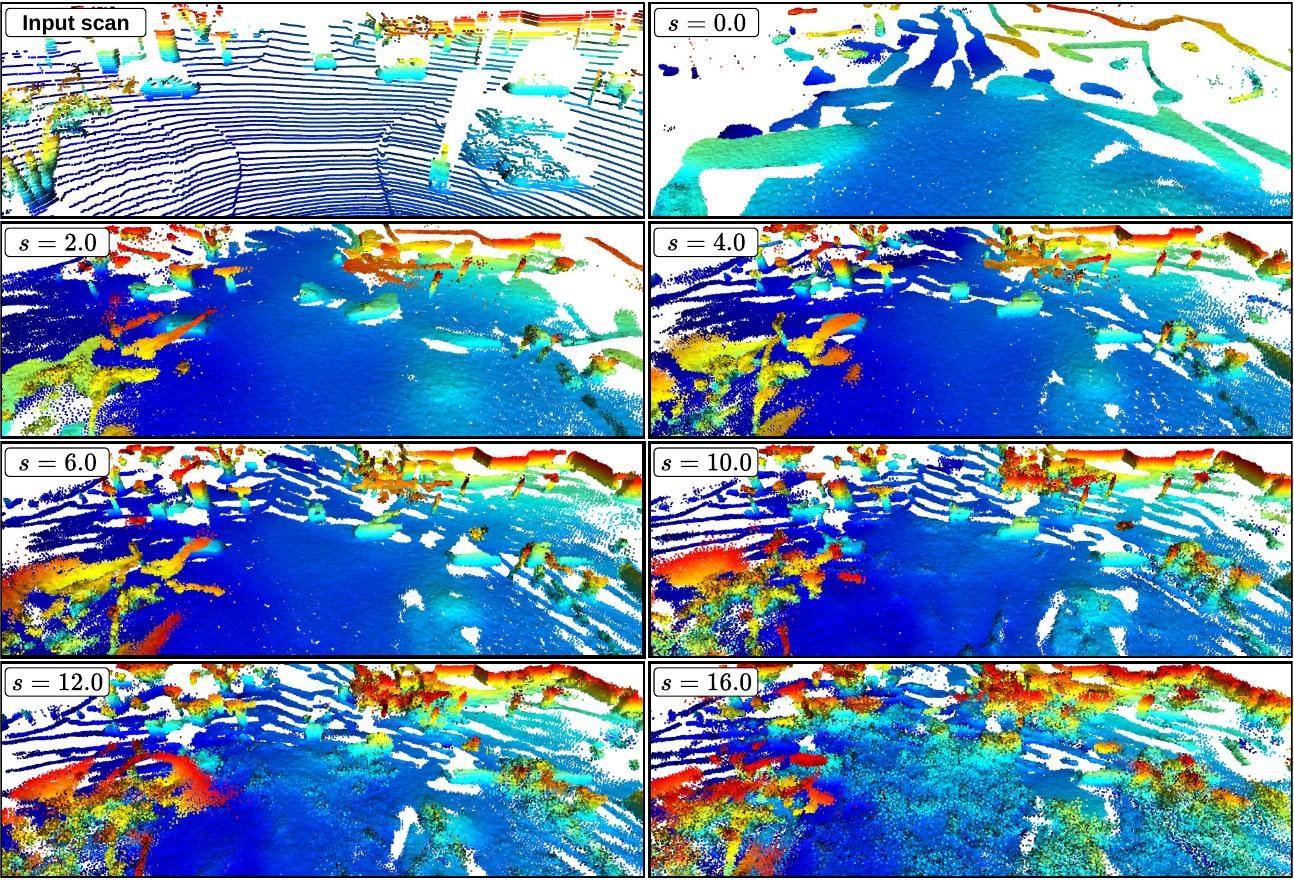}
     \caption{Comparison between results with different conditioning weights $s$.}
     \label{fig:cond_weights}
\end{figure*}

\begin{figure*}[t]
    \centering
    \includegraphics[width=1.0\linewidth]{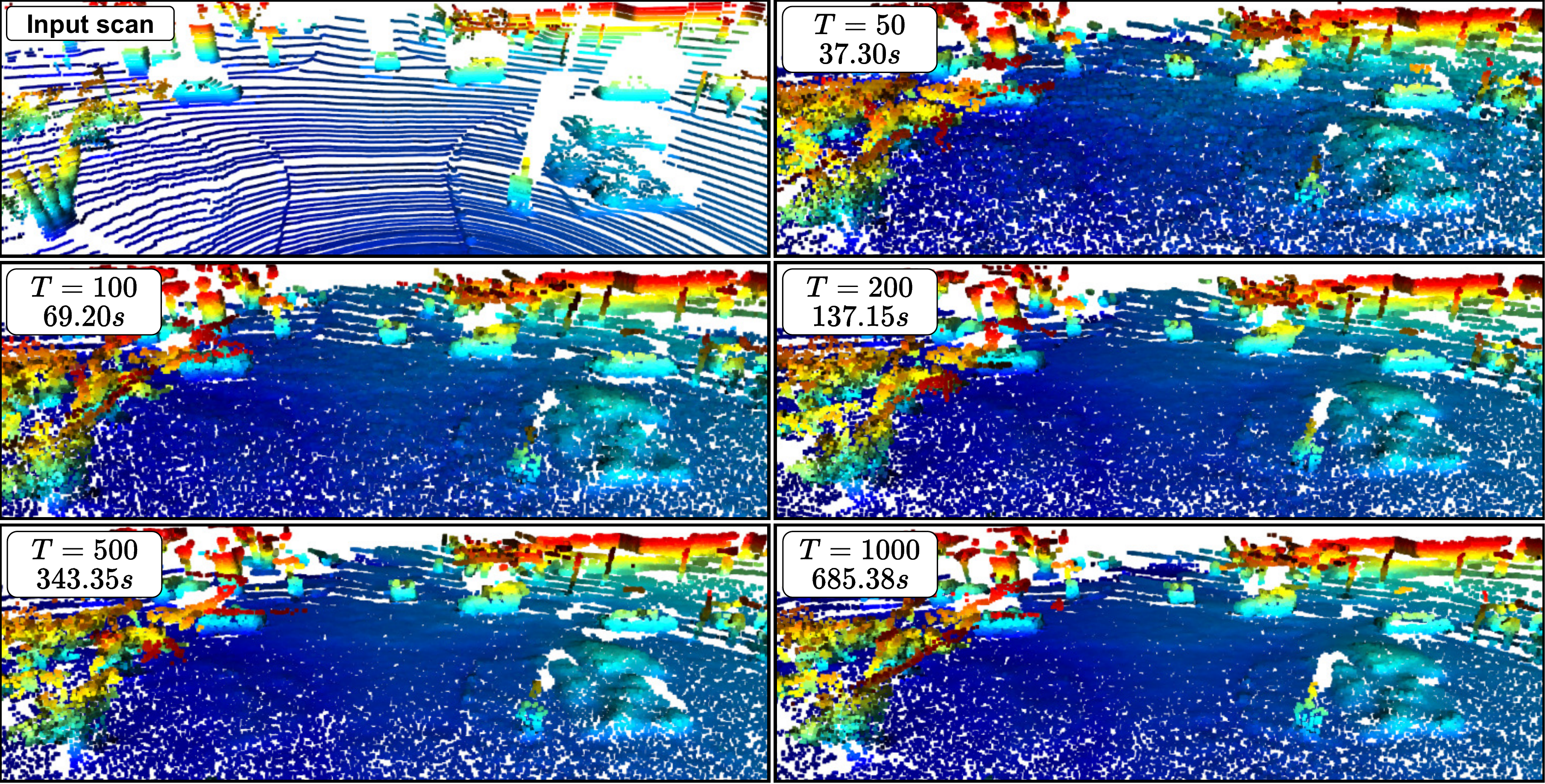}
     \caption{Comparison between results with different number of denoising steps $T$.}
     \label{fig:denoising_steps}
\end{figure*}

This section compares the scene completion quality using different condition weights $s$ qualitatively and quantitatively. \cref{fig:cond_weights}
shows the qualitative comparison between the scene completion with different conditioning weights. With $s=0.0$, we have the unconditional generation.
In this case, the generated scene has a flat surface distributed over the input scan borders without retaining structural information.
As we increase $s$, the structure details are better defined. With $s=2.0$ and $s=4.0$ more details are generated but with a smooth aspect.
With $s = 6.0$ the generation follows structural information from the input scan and defines sharper boundaries over the structures.
With $s=10.0$ and $s=12.0$, the generated scene gets too noisy, generating artifacts over the scene.

We also evaluate the influence of the conditioning weight~$s$ in \cref{tab:chamfer_dist_cond_weights}. As in Tab. 5 of the main paper,
we compute the chamfer distance over the scene completion and the ground truth over a short sequence from the SemanticKITTI validation set, where
we generate every one hundred scans. In this evaluation, having $s=6.0$ and $s=10.0$ achieves basically the same performance. However,
from the qualitative evaluation presented in \cref{fig:cond_weights}, we used $s=6.0$ in the main paper since it achieved the best performance visually
and numerically.

\begin{table}
    \centering
    \small
    \setlength{\tabcolsep}{0.15cm}
    \begin{tabular}{lccccccc}
      \toprule 
      s & 0.0 & 2.0 & 4.0 & 6.0 & 10.0 & 12.0 & 16.0 \\
      \midrule
      CD [m] & 0.737 & 0.543 & 0.454 & \textbf{0.433} & \textbf{0.432} & 0.435 & 0.450 \\
      \bottomrule
    \end{tabular}
    \caption{Mean chamfer distance over a short sequence from the validation set of SemanticKITTI with different conditioning weights $s$.}
    \label{tab:chamfer_dist_cond_weights}
    \vspace{-0.35cm}
  \end{table}

\section{Denoising steps}
\label{sec:denoising_steps}

In this section, we compare the quality of the scene completion with different number of denoising steps $T$. \cref{fig:denoising_steps} shows the diffusion generation
using DPMSolver~\cite{lu2022neurips} with the different number of denoising steps and the amount of time in seconds to generate the complete scene. Since the model
was trained with $T=1,000$, we can achieve the best quality result when using $T=1,000$ during inference. However, inferring the $1,000$ steps demands many computational
time. As we decrease $T$, we increase the inference speed. However, we can also notice that with lower $T$,
the scene generation loses details. This can be seen when comparing the structures in the scene, especially the ground, where more noise can be noticed
as we decrease $T$. Therefore, in the main paper we set $T=50$ and take advantage of the refinement network to compensate for the lower quality generation
when using smaller $T$.

\section{Further qualitative results}
\label{sec:qual_res}

\begin{figure*}[t]
    \centering
    \includegraphics[width=1.0\linewidth]{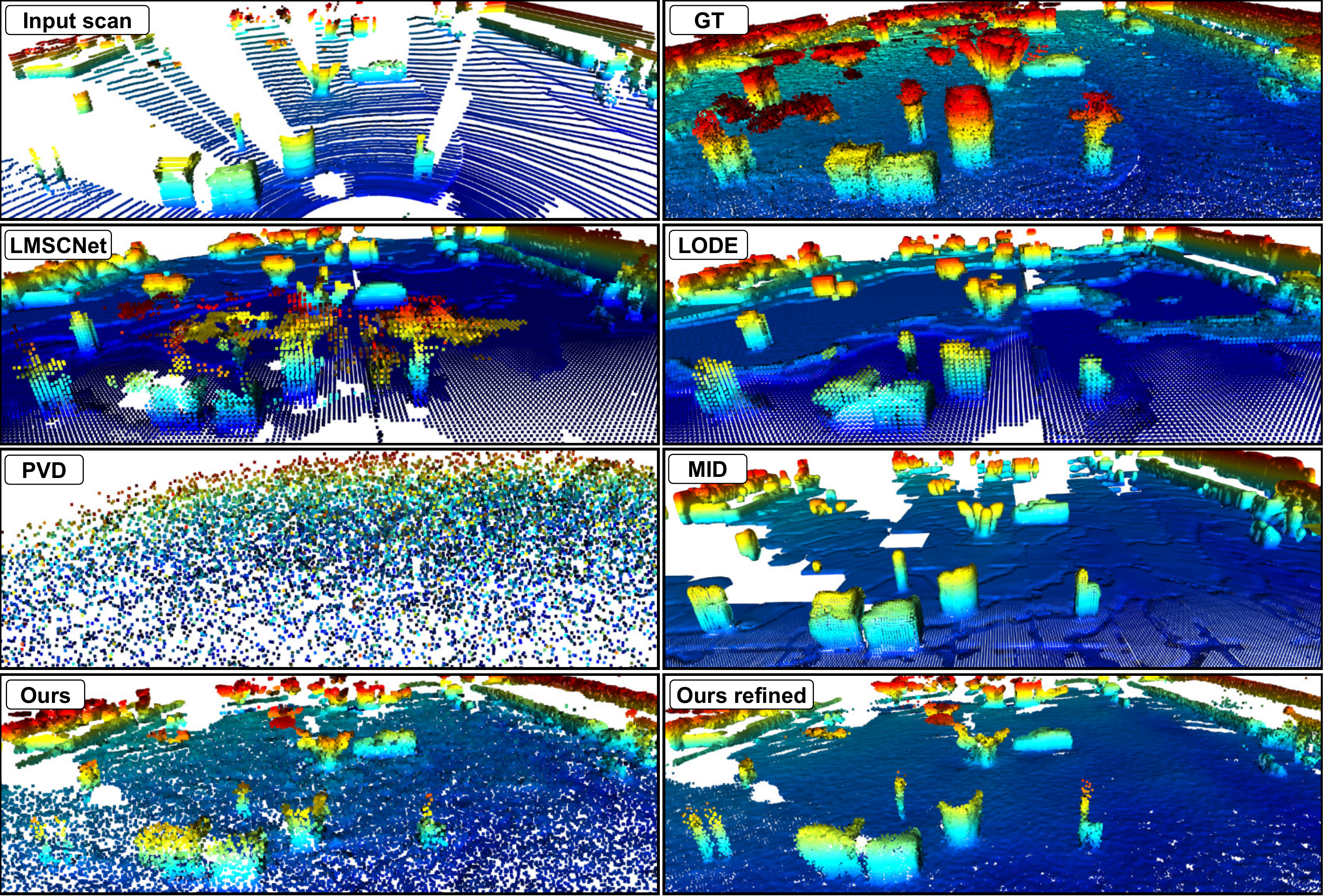}
     \caption{Qualitative results comparing the scene completion between our method and the baselines evaluated in the main paper.}
     \label{fig:qualitative_1}
\end{figure*}

\begin{figure*}[t]
    \centering
    \includegraphics[width=1.0\linewidth]{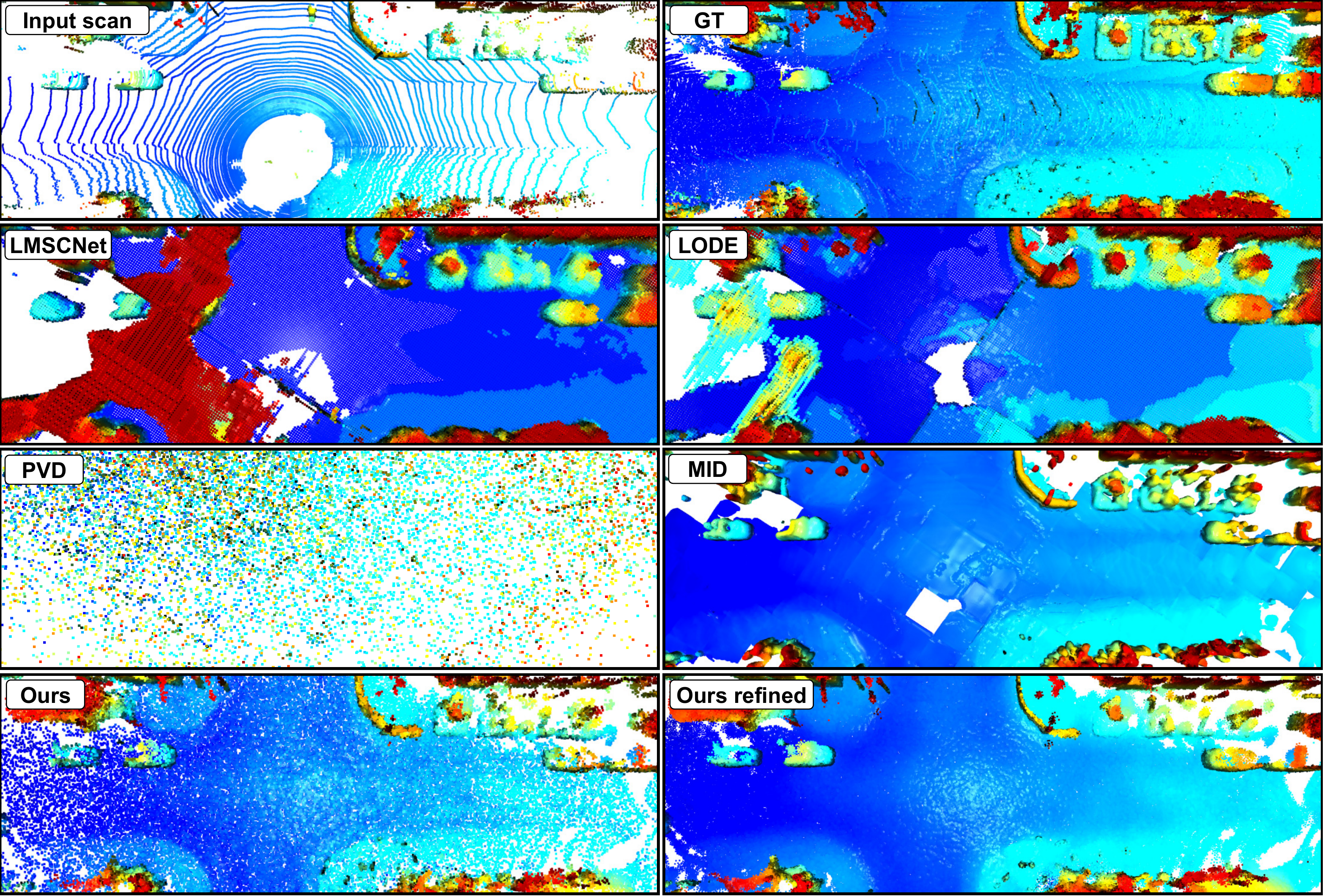}
     \caption{Qualitative results comparing the scene completion between our method and the baselines evaluated in the main paper.}
     \label{fig:qualitative_2}
\end{figure*}

\begin{figure*}[t]
    \centering
    \includegraphics[width=1.0\linewidth]{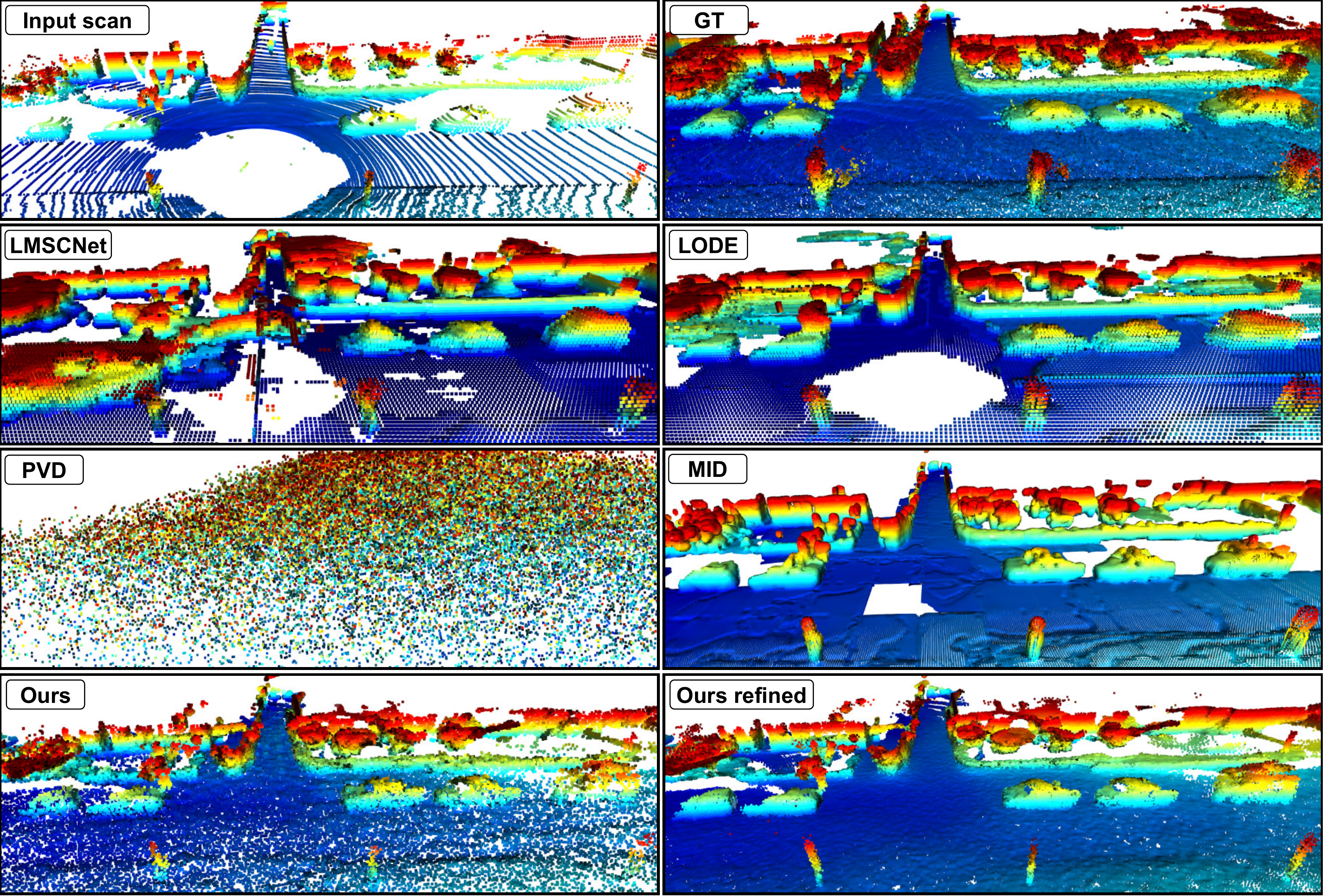}
     \caption{Qualitative results comparing the scene completion between our method and the baselines evaluated in the main paper.}
     \label{fig:qualitative_3}
\end{figure*}

\begin{figure*}[t]
    \centering
    \includegraphics[width=1.0\linewidth]{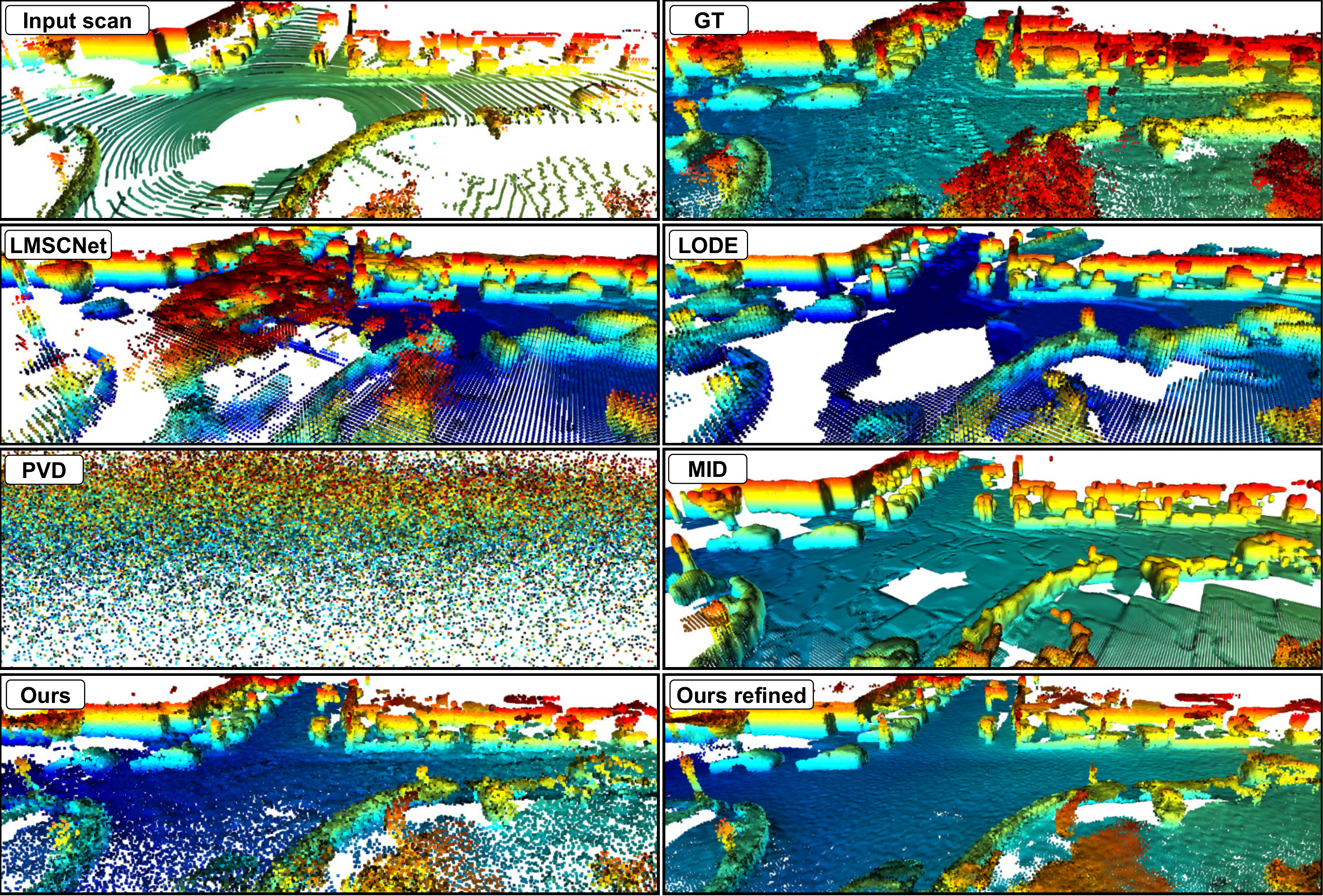}
     \caption{Qualitative results comparing the scene completion between our method and the baselines evaluated in the main paper.}
     \label{fig:qualitative_4}
\end{figure*}

\begin{figure*}[t]
    \centering
    \includegraphics[width=1.0\linewidth]{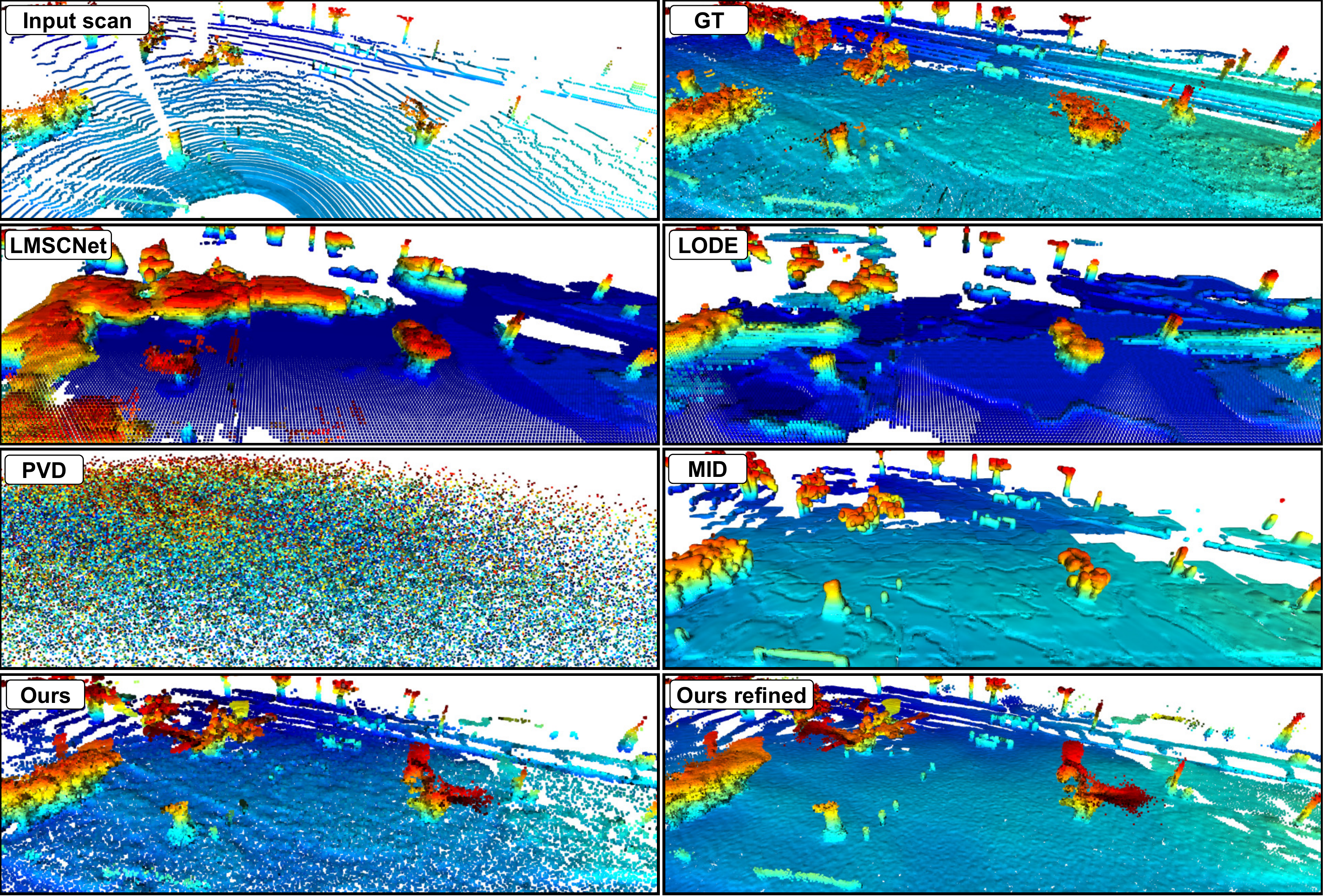}
     \caption{Qualitative results comparing the scene completion between our method and the baselines evaluated in the main paper.}
     \label{fig:qualitative_5}
\end{figure*}

In this section, we show more qualitative results, comparing our scene completion with the baselines evaluated in the paper, \ie,  LMSCNet~\cite{roldao2020threedv}, PVD~\cite{zhou2021iccv-sgac}, Make It Dense (MID)~\cite{vizzo2022ral}, and LODE~\cite{li2023icra-llce}.
\cref{fig:qualitative_1,fig:qualitative_2,fig:qualitative_3,fig:qualitative_4,fig:qualitative_5} compare the results between the baselines and our method.
As shown, the diffusion baseline PVD~\cite{zhou2021iccv-sgac} fails to generate scene-scale data. The SDF baselines reconstruct the scene inheriting
artifacts from the surface approximation and the voxelization. Our method achieves a more detailed representation, with a smoother generation compared to the baselines.

{
    \small
    \bibliographystyle{ieeenat_fullname}
    \bibliography{glorified,new}
}

%% file: preamble.tex
\usepackage[dvipsnames]{xcolor}

%% file: nunes2024cvpr.bbl
\begin{thebibliography}{50}
\providecommand{\natexlab}[1]{#1}
\providecommand{\url}[1]{\texttt{#1}}
\expandafter\ifx\csname urlstyle\endcsname\relax
  \providecommand{\doi}[1]{doi: #1}\else
  \providecommand{\doi}{doi: \begingroup \urlstyle{rm}\Url}\fi

\bibitem[Balaji et~al.(2022)Balaji, Nah, Huang, Vahdat, Song, Kreis, Aittala,
  Aila, Laine, Catanzaro, Karras, and Liu]{balaji2022arxiv}
Yogesh Balaji, Seungjun Nah, Xun Huang, Arash Vahdat, Jiaming Song, Karsten
  Kreis, Miika Aittala, Timo Aila, Samuli Laine, Bryan Catanzaro, Tero Karras,
  and Ming-Yu Liu.
\newblock ediff-i: Text-to-image diffusion models with an ensemble of expert
  denoisers.
\newblock \emph{arXiv preprint}, arXiv:2211.01324, 2022.

\bibitem[Behley et~al.(2019)Behley, Garbade, Milioto, Quenzel, Behnke,
  Stachniss, and Gall]{behley2019iccv}
Jens Behley, Martin Garbade, Andres Milioto, Jan Quenzel, Sven Behnke, Cyrill
  Stachniss, and Juergen Gall.
\newblock {SemanticKITTI: A Dataset for Semantic Scene Understanding of LiDAR
  Sequences}.
\newblock In \emph{Proc.~of the IEEE/CVF Intl.~Conf.~on Computer Vision
  (ICCV)}, 2019.

\bibitem[Behley et~al.(2021)Behley, Garbade, Milioto, Quenzel, Behnke, Gall,
  and Stachniss]{behley2021ijrr}
Jens Behley, Martin Garbade, Andres Milioto, Jan Quenzel, Sven Behnke, Juergen
  Gall, and Cyrill Stachniss.
\newblock {Towards 3D LiDAR-based Semantic Scene Understanding of 3D Point
  Cloud Sequences: The SemanticKITTI Dataset}.
\newblock \emph{Intl.~Journal~of Robotics Research (IJRR)}, 40\penalty0
  (8--9):\penalty0 959--967, 2021.

\bibitem[Caesar et~al.(2020)Caesar, Bankiti, Lang, Vora, Liong, Xu, Krishnan,
  Pan, Baldan, and Beijbom]{caesar2020cvpr}
Holger Caesar, Varun Bankiti, Alex~H. Lang, Sourabh Vora, Venice~Erin Liong,
  Qiang Xu, Anush Krishnan, Yu Pan, Giancarlo Baldan, and Oscar Beijbom.
\newblock {nuScenes: A Multimodal Dataset for Autonomous Driving}.
\newblock In \emph{Proc.~of the IEEE/CVF Conf.~on Computer Vision and Pattern
  Recognition (CVPR)}, 2020.

\bibitem[Choy et~al.(2019)Choy, Gwak, and Savarese]{choy2019cvpr}
Christopher Choy, JunYoung Gwak, and Silvio Savarese.
\newblock {4D Spatio-Temporal ConvNets: Minkowski Convolutional Neural
  Networks}.
\newblock In \emph{Proc.~of the IEEE/CVF Conf.~on Computer Vision and Pattern
  Recognition (CVPR)}, 2019.

\bibitem[Dhariwal and Nichol(2021)]{dhariwal2021neurips}
Prafulla Dhariwal and Alexander Nichol.
\newblock Diffusion models beat gans on image synthesis.
\newblock In \emph{Proc.~of the Conf. on Neural Information Processing Systems
  (NeurIPS)}, 2021.

\bibitem[Fong et~al.(2022)Fong, Mohan, Hurtado, Zhou, Caesar, Beijbom, and
  Valada]{fong2022icra}
Whye~Kit Fong, Rohit Mohan, Juana~Valeria Hurtado, Lubing Zhou, Holgr Caesar,
  Oscar Beijbom, and Abhinav Valada.
\newblock {Panoptic nuScenes A Large-Scale Benchmark for LiDAR Panoptic
  Segmentation and Tracking}.
\newblock In \emph{Proc.~of the IEEE Intl.~Conf.~on Robotics \& Automation
  (ICRA)}, 2022.

\bibitem[Fu et~al.(2020)Fu, Dong, Mertz, and Dolan]{fu2020iros-dcvi}
Chen Fu, Chiyu Dong, Christoph Mertz, and John~M. Dolan.
\newblock {Depth Completion Via Inductive Fusion of Planar LIDAR and Monocular
  Camera}.
\newblock In \emph{Proc.~of the IEEE/RSJ Intl.~Conf.~on Intelligent Robots and
  Systems (IROS)}, 2020.

\bibitem[Geiger et~al.(2012)Geiger, Lenz, and Urtasun]{geiger2012cvpr}
Andreas Geiger, Philip Lenz, and Raquel Urtasun.
\newblock {Are we ready for Autonomous Driving? The KITTI Vision Benchmark
  Suite}.
\newblock In \emph{Proc.~of the IEEE Conf.~on Computer Vision and Pattern
  Recognition (CVPR)}, 2012.

\bibitem[Ho and Salimans(2021)]{ho2021neuripsws}
Jonathan Ho and Tim Salimans.
\newblock Classifier-free diffusion guidance.
\newblock In \emph{NeurIPS 2021 Workshop on Deep Generative Models and
  Downstream Applications}, 2021.

\bibitem[Ho et~al.(2020)Ho, Jain, and Abbeel]{ho2020neurips}
Jonathan Ho, Ajay Jain, and Pieter Abbeel.
\newblock Denoising diffusion probabilistic models.
\newblock In \emph{Proc.~of the Conf. on Neural Information Processing Systems
  (NeurIPS)}, 2020.

\bibitem[Karras et~al.(2022)Karras, Aittala, Aila, and
  Laine]{karras2022neurips}
Tero Karras, Miika Aittala, Timo Aila, and Samuli Laine.
\newblock Elucidating the design space of diffusion-based generative models.
\newblock In \emph{Proc.~of the Conf. on Neural Information Processing Systems
  (NeurIPS)}, 2022.

\bibitem[Kingma and Ba(2015)]{kingma2015iclr}
Diederik~P. Kingma and Jimmy Ba.
\newblock {Adam: {A} Method for Stochastic Optimization}.
\newblock In \emph{Proc.~of the Int.~Conf.~on Learning Representations (ICLR)},
  2015.

\bibitem[Lee et~al.(2023)Lee, Im, Lee, and Yoon]{lee2023arxiv}
Jumin Lee, Woobin Im, Sebin Lee, and Sung-Eui Yoon.
\newblock Diffusion probabilistic models for scene-scale 3d categorical data.
\newblock \emph{arXiv preprint}, arXiv:2301.00527, 2023.

\bibitem[Li et~al.(2023)Li, Zhao, Shi, Zhao, Yuan, Zhou, and
  Zhang]{li2023icra-llce}
Pengfei Li, Ruowen Zhao, Yongliang Shi, Hao Zhao, Jirui Yuan, Guyue Zhou, and
  Ya-Qin Zhang.
\newblock {LODE Locally Conditioned Eikonal Implicit Scene Completion from
  Sparse LiDAR}.
\newblock In \emph{Proc.~of the IEEE Intl.~Conf.~on Robotics \& Automation
  (ICRA)}, 2023.

\bibitem[Liao et~al.(2022)Liao, Xie, and Geiger]{liao2022pami}
Yiyi Liao, Jun Xie, and Andreas Geiger.
\newblock {KITTI}-360: A novel dataset and benchmarks for urban scene
  understanding in 2d and 3d.
\newblock \emph{IEEE Trans.~on Pattern Analysis and Machine Intelligence
  (TPAMI)}, 2022.

\bibitem[Lu et~al.(2022)Lu, Zhou, Bao, Chen, Li, and Zhu]{lu2022neurips}
Cheng Lu, Yuhao Zhou, Fan Bao, Jianfei Chen, Chongxuan Li, and Jun Zhu.
\newblock {DPM}-solver: A fast {ODE} solver for diffusion probabilistic model
  sampling in around 10 steps.
\newblock In \emph{Proc.~of the Conf. on Neural Information Processing Systems
  (NeurIPS)}, 2022.

\bibitem[Lu et~al.(2023)Lu, Zhou, Bao, Chen, Li, and Zhu]{lu2023arxiv}
Cheng Lu, Yuhao Zhou, Fan Bao, Jianfei Chen, Chongxuan Li, and Jun Zhu.
\newblock Dpm-solver++: Fast solver for guided sampling of diffusion
  probabilistic models.
\newblock \emph{arXiv preprint}, arXiv:2211.01095, 2023.

\bibitem[Luo and Hu(2021)]{luo2021cvpr}
Shitong Luo and Wei Hu.
\newblock Diffusion probabilistic models for 3d point cloud generation.
\newblock In \emph{Proc.~of the IEEE/CVF Conf.~on Computer Vision and Pattern
  Recognition (CVPR)}, 2021.

\bibitem[Lyu et~al.(2022)Lyu, Kong, XU, Pan, and Lin]{lyu2022iclr}
Zhaoyang Lyu, Zhifeng Kong, Xudong XU, Liang Pan, and Dahua Lin.
\newblock A conditional point diffusion-refinement paradigm for 3d point cloud
  completion.
\newblock In \emph{Proc.~of the Int.~Conf.~on Learning Representations (ICLR)},
  2022.

\bibitem[Ma and Karaman(2018)]{ma2018icra}
Fangchang Ma and Sertac Karaman.
\newblock {Sparse-To-Dense: Depth Prediction from Sparse Depth Samples and a
  Single Image}.
\newblock In \emph{Proc.~of the IEEE Intl.~Conf.~on Robotics \& Automation
  (ICRA)}, 2018.

\bibitem[Ma et~al.(2019)Ma, Cavalheiro, and Karaman]{ma2019icra-sssdcf}
Fangchang Ma, Guilherme~Venturelli Cavalheiro, and Sertac Karaman.
\newblock {Self-Supervised Sparse-To-Dense Self-Supervised Depth Completion
  from LiDAR and Monocular Camera}.
\newblock In \emph{Proc.~of the IEEE Intl.~Conf.~on Robotics \& Automation
  (ICRA)}, 2019.

\bibitem[Meng et~al.(2023)Meng, Rombach, Gao, Kingma, Ermon, Ho, and
  Salimans]{meng2023cvpr}
Chenlin Meng, Robin Rombach, Ruiqi Gao, Diederik Kingma, Stefano Ermon,
  Jonathan Ho, and Tim Salimans.
\newblock On distillation of guided diffusion models.
\newblock In \emph{Proc.~of the IEEE/CVF Conf.~on Computer Vision and Pattern
  Recognition (CVPR)}, 2023.

\bibitem[Mersch et~al.(2023)Mersch, Guadagnino, Chen, Tiziano, Vizzo, Behley,
  and Stachniss]{mersch2023ral}
Benedikt Mersch, Tiziano Guadagnino, Xieyuanli Chen, Tiziano, Ignacio Vizzo,
  Jens Behley, and Cyrill Stachniss.
\newblock {Building Volumetric Beliefs for Dynamic Environments Exploiting
  Map-Based Moving Object Segmentation}.
\newblock \emph{IEEE Robotics and Automation Letters (RA-L)}, 8\penalty0
  (8):\penalty0 5180--5187, 2023.

\bibitem[Murez et~al.(2020)Murez, van As, Bartolozzi, Sinha, Badrinarayanan,
  and Rabinovich]{murez2020eccv}
Zak Murez, Tarrence van As, James Bartolozzi, Ayan Sinha, Vijay Badrinarayanan,
  and Andrew Rabinovich.
\newblock Atlas: End-to-end 3d scene reconstruction from posed images.
\newblock In \emph{Proc.~of the Europ.~Conf.~on Computer Vision (ECCV)}, 2020.

\bibitem[Nakashima and Kurazume(2023)]{nakashima2023arxiv}
Kazuto Nakashima and Ryo Kurazume.
\newblock Lidar data synthesis with denoising diffusion probabilistic models.
\newblock \emph{arXiv preprint}, arXiV:2309.09256, 2023.

\bibitem[Nichol and Dhariwal(2021)]{nichol2021pmlr}
Alexander~Quinn Nichol and Prafulla Dhariwal.
\newblock Improved denoising diffusion probabilistic models.
\newblock In \emph{Proc.~of Machine Learning Research (PMLR)}, 2021.

\bibitem[Peebles and Xie(2023)]{peebles2023iccv}
William Peebles and Saining Xie.
\newblock Scalable diffusion models with transformers.
\newblock In \emph{Proc.~of the IEEE/CVF Intl.~Conf.~on Computer Vision
  (ICCV)}, 2023.

\bibitem[Popović et~al.(2021)Popović, Thomas, Papatheodorou, Funk,
  Vidal-Calleja, and Leutenegger]{popovic2021ral}
Marija Popović, Florian Thomas, Sotiris Papatheodorou, Nils Funk, Teresa
  Vidal-Calleja, and Stefan Leutenegger.
\newblock {Volumetric Occupancy Mapping With Probabilistic Depth Completion for
  Robotic Navigation}.
\newblock \emph{IEEE Robotics and Automation Letters (RA-L)}, 6\penalty0
  (3):\penalty0 5072--5079, 2021.

\bibitem[Ramesh et~al.(2021)Ramesh, Pavlov, Goh, Gray, Voss, Radford, Chen, and
  Sutskever]{ramesh2021pmlr}
Aditya Ramesh, Mikhail Pavlov, Gabriel Goh, Scott Gray, Chelsea Voss, Alec
  Radford, Mark Chen, and Ilya Sutskever.
\newblock Zero-shot text-to-image generation.
\newblock In \emph{Proc.~of the International Conference on Machine Learning},
  2021.

\bibitem[Rist et~al.(2021)Rist, Emmerichs, Enzweiler, and
  Gavrila]{rist2021pmai}
Christoph Rist, David Emmerichs, Markus Enzweiler, and Dariu Gavrila.
\newblock Semantic scene completion using local deep implicit functions on
  lidar data.
\newblock \emph{IEEE Trans.~on Pattern Analysis and Machine Intelligence
  (TPAMI)}, 44\penalty0 (10):\penalty0 7205--7218, 2021.

\bibitem[Rold{\~a}o et~al.(2020)Rold{\~a}o, de~Charette, and
  Verroust-Blondet]{roldao2020threedv}
Luis Rold{\~a}o, Raoul de Charette, and Anne Verroust-Blondet.
\newblock {LMSCNet: Lightweight Multiscale 3D Semantic Completion}.
\newblock In \emph{Proc.~of the Intl.~Conf.~on 3D Vision (3DV)}, 2020.

\bibitem[Rombach et~al.(2022)Rombach, Blattmann, Lorenz, Esser, and
  Ommer]{rombach2022cvpr}
Robin Rombach, Andreas Blattmann, Dominik Lorenz, Patrick Esser, and Bj\"orn
  Ommer.
\newblock {High-Resolution Image Synthesis With Latent Diffusion Models}.
\newblock In \emph{Proc.~of the IEEE/CVF Conf.~on Computer Vision and Pattern
  Recognition (CVPR)}, 2022.

\bibitem[Salimans and Ho(2022)]{salimans2022iclr}
Tim Salimans and Jonathan Ho.
\newblock Progressive distillation for fast sampling of diffusion models.
\newblock In \emph{Proc.~of the Int.~Conf.~on Learning Representations (ICLR)},
  2022.

\bibitem[Sanghi et~al.(2022)Sanghi, Chu, Lambourne, Wang, Cheng, Fumero, and
  Malekshan]{sanghi2022cvpr}
Aditya Sanghi, Hang Chu, Joseph~G. Lambourne, Ye Wang, Chin-Yi Cheng, Marco
  Fumero, and Kamal~Rahimi Malekshan.
\newblock {CLIP-Forge: Towards Zero-Shot Text-To-Shape Generation}.
\newblock In \emph{Proc.~of the IEEE/CVF Conf.~on Computer Vision and Pattern
  Recognition (CVPR)}, 2022.

\bibitem[Sanghi et~al.(2023)Sanghi, Fu, Liu, Willis, Shayani, Khasahmadi,
  Sridhar, and Ritchie]{sanghi2023cvpr}
Aditya Sanghi, Rao Fu, Vivian Liu, Karl~D.D. Willis, Hooman Shayani, Amir~H.
  Khasahmadi, Srinath Sridhar, and Daniel Ritchie.
\newblock {CLIP-Sculptor: Zero-Shot Generation of High-Fidelity and Diverse
  Shapes From Natural Language}.
\newblock In \emph{Proc.~of the IEEE/CVF Conf.~on Computer Vision and Pattern
  Recognition (CVPR)}, 2023.

\bibitem[Song et~al.(2021)Song, Meng, and Ermon]{song2021iclr}
Jiaming Song, Chenlin Meng, and Stefano Ermon.
\newblock Denoising diffusion implicit models.
\newblock In \emph{Proc.~of the Int.~Conf.~on Learning Representations (ICLR)},
  2021.

\bibitem[Song et~al.(2017)Song, Yu, Zeng, Chang, Savva, and
  Funkhouser]{song2017cvpr}
Shuran Song, Fisher Yu, Andy Zeng, Angel~X. Chang, Manolis Savva, and Thomas
  Funkhouser.
\newblock {Semantic Scene Completion from a Single Depth Image}.
\newblock In \emph{Proc.~of the IEEE Conf.~on Computer Vision and Pattern
  Recognition (CVPR)}, 2017.

\bibitem[Sun et~al.(2020)Sun, Kretzschmar, Dotiwalla, Chouard, Patnaik, Tsui,
  Guo, Zhou, Chai, Caine, Vasudevan, Han, Ngiam, Zhao, Timofeev, Ettinger,
  Krivokon, Gao, Joshi, Zhao, Cheng, Zhang, Shlens, Chen, and
  Anguelov]{sun2020cvpr}
Pei Sun, Henrik Kretzschmar, Xerxes Dotiwalla, Aurelien Chouard, Vijaysai
  Patnaik, Paul Tsui, James Guo, Yin Zhou, Yuning Chai, Benjamin Caine, Vijay
  Vasudevan, Wei Han, Jiquan Ngiam, Hang Zhao, Aleksei Timofeev, Scott
  Ettinger, Maxim Krivokon, Amy Gao, Aditya Joshi, Sheng Zhao, Shuyang Cheng,
  Yu Zhang, Jonathon Shlens, Zhifeng Chen, and Dragomir Anguelov.
\newblock Scalability in perception for autonomous driving: Waymo open dataset.
\newblock In \emph{Proc.~of the IEEE/CVF Conf.~on Computer Vision and Pattern
  Recognition (CVPR)}, 2020.

\bibitem[Vizzo et~al.(2022)Vizzo, Mersch, Marcuzzi, Wiesmann, Behley, and
  Stachniss]{vizzo2022ral}
Ignacio Vizzo, Benedikt Mersch, Rodrigo Marcuzzi, Louis Wiesmann, Jens Behley,
  and Cyrill Stachniss.
\newblock Make it dense: Self-supervised geometric scan completion of sparse 3d
  lidar scans in large outdoor environments.
\newblock \emph{IEEE Robotics and Automation Letters (RA-L)}, 7\penalty0
  (3):\penalty0 8534--8541, 2022.

\bibitem[Vizzo et~al.(2023)Vizzo, Guadagnino, Mersch, Wiesmann, Behley, and
  Stachniss]{vizzo2023ral}
Ignacio Vizzo, Tiziano Guadagnino, Benedikt Mersch, Louis Wiesmann, Jens
  Behley, and Cyrill Stachniss.
\newblock {KISS-ICP: In Defense of Point-to-Point ICP -- Simple, Accurate, and
  Robust Registration If Done the Right Way}.
\newblock \emph{IEEE Robotics and Automation Letters (RA-L)}, 8\penalty0
  (2):\penalty0 1029--1036, 2023.

\bibitem[Xiong et~al.(2023)Xiong, Ma, Wang, and Urtasun]{xiong2023cvpr-lcrf}
Yuwen Xiong, Wei-Chiu Ma, Jingkang Wang, and Raquel Urtasun.
\newblock {Learning Compact Representations for LiDAR Completion and
  Generation}.
\newblock In \emph{Proc.~of the IEEE/CVF Conf.~on Computer Vision and Pattern
  Recognition (CVPR)}, 2023.

\bibitem[Xu et~al.(2023)Xu, Wang, Cheng, Cao, Shan, Qie, and
  Gao]{xu2023cvpr-dzts}
Jiale Xu, Xintao Wang, Weihao Cheng, Yan-Pei Cao, Ying Shan, Xiaohu Qie, and
  Shenghua Gao.
\newblock {Dream3D: Zero-Shot Text-to-3D Synthesis Using 3D Shape Prior and
  Text-to-Image Diffusion Models}.
\newblock In \emph{Proc.~of the IEEE/CVF Conf.~on Computer Vision and Pattern
  Recognition (CVPR)}, 2023.

\bibitem[Xu et~al.(2019)Xu, Zhu, Shi, Zhang, Bao, and Li]{xu2019iccv-dcfs}
Yan Xu, Xinge Zhu, Jianping Shi, Guofeng Zhang, Hujun Bao, and Hongsheng Li.
\newblock {Depth Completion From Sparse LiDAR Data With Depth-Normal
  Constraints}.
\newblock In \emph{Proc.~of the IEEE/CVF Intl.~Conf.~on Computer Vision
  (ICCV)}, 2019.

\bibitem[Zeng et~al.(2022)Zeng, Vahdat, Williams, Gojcic, Litany, Fidler, and
  Kreis]{zeng2022neurips}
Xiaohui Zeng, Arash Vahdat, Francis Williams, Zan Gojcic, Or Litany, Sanja
  Fidler, and Karsten Kreis.
\newblock Lion: Latent point diffusion models for 3d shape generation.
\newblock In \emph{Proc.~of the Conf. on Neural Information Processing Systems
  (NeurIPS)}, 2022.

\bibitem[Zhang et~al.(2023)Zhang, Rao, and Agrawala]{zhang2023iccv}
Lvmin Zhang, Anyi Rao, and Maneesh Agrawala.
\newblock Adding conditional control to text-to-image diffusion models.
\newblock In \emph{Proc.~of the IEEE/CVF Intl.~Conf.~on Computer Vision
  (ICCV)}, 2023.

\bibitem[Zhou et~al.(2021)Zhou, Du, and Wu]{zhou2021iccv-sgac}
Linqi Zhou, Yilun Du, and Jiajun Wu.
\newblock {3D Shape Generation and Completion Through Point-Voxel Diffusion}.
\newblock In \emph{Proc.~of the IEEE/CVF Intl.~Conf.~on Computer Vision
  (ICCV)}, 2021.

\bibitem[Zhou et~al.(2022)Zhou, Zhang, Chen, Li, Tensmeyer, Yu, Gu, Xu, and
  Sun]{zhou2022cvpr-tltf}
Yufan Zhou, Ruiyi Zhang, Changyou Chen, Chunyuan Li, Chris Tensmeyer, Tong Yu,
  Jiuxiang Gu, Jinhui Xu, and Tong Sun.
\newblock {Towards Language-Free Training for Text-to-Image Generation}.
\newblock In \emph{Proc.~of the IEEE/CVF Conf.~on Computer Vision and Pattern
  Recognition (CVPR)}, 2022.

\bibitem[Zhou et~al.(2023)Zhou, Liu, Zhu, Yang, Chen, and
  Xu]{zhou2023cvpr-sdft}
Yufan Zhou, Bingchen Liu, Yizhe Zhu, Xiao Yang, Changyou Chen, and Jinhui Xu.
\newblock {Shifted Diffusion for Text-to-Image Generation}.
\newblock In \emph{Proc.~of the IEEE/CVF Conf.~on Computer Vision and Pattern
  Recognition (CVPR)}, 2023.

\bibitem[Zyrianov et~al.(2022)Zyrianov, Zhu, and Wang]{zyrianov2022eccv}
Vlas Zyrianov, Xiyue Zhu, and Shenlong Wang.
\newblock Learning to generate realistic lidar point clouds.
\newblock In \emph{Proc.~of the Europ.~Conf.~on Computer Vision (ECCV)}, 2022.

\end{thebibliography}


\begin{thebibliography}{8}
\providecommand{\natexlab}[1]{#1}
\providecommand{\url}[1]{\texttt{#1}}
\expandafter\ifx\csname urlstyle\endcsname\relax
  \providecommand{\doi}[1]{doi: #1}\else
  \providecommand{\doi}{doi: \begingroup \urlstyle{rm}\Url}\fi

\bibitem[Choy et~al.(2019)Choy, Gwak, and Savarese]{choy2019cvpr}
Christopher Choy, JunYoung Gwak, and Silvio Savarese.
\newblock {4D Spatio-Temporal ConvNets: Minkowski Convolutional Neural
  Networks}.
\newblock In \emph{Proc.~of the IEEE/CVF Conf.~on Computer Vision and Pattern
  Recognition (CVPR)}, 2019.

\bibitem[Kingma and Ba(2015)]{kingma2015iclr}
Diederik~P. Kingma and Jimmy Ba.
\newblock {Adam: {A} Method for Stochastic Optimization}.
\newblock In \emph{Proc.~of the Int.~Conf.~on Learning Representations (ICLR)},
  2015.

\bibitem[Li et~al.(2023)Li, Zhao, Shi, Zhao, Yuan, Zhou, and
  Zhang]{li2023icra-llce}
Pengfei Li, Ruowen Zhao, Yongliang Shi, Hao Zhao, Jirui Yuan, Guyue Zhou, and
  Ya-Qin Zhang.
\newblock {LODE Locally Conditioned Eikonal Implicit Scene Completion from
  Sparse LiDAR}.
\newblock In \emph{Proc.~of the IEEE Intl.~Conf.~on Robotics \& Automation
  (ICRA)}, 2023.

\bibitem[Lu et~al.(2022)Lu, Zhou, Bao, Chen, Li, and Zhu]{lu2022neurips}
Cheng Lu, Yuhao Zhou, Fan Bao, Jianfei Chen, Chongxuan Li, and Jun Zhu.
\newblock {DPM}-solver: A fast {ODE} solver for diffusion probabilistic model
  sampling in around 10 steps.
\newblock In \emph{Proc.~of the Conf. on Neural Information Processing Systems
  (NeurIPS)}, 2022.

\bibitem[Lyu et~al.(2022)Lyu, Kong, XU, Pan, and Lin]{lyu2022iclr}
Zhaoyang Lyu, Zhifeng Kong, Xudong XU, Liang Pan, and Dahua Lin.
\newblock A conditional point diffusion-refinement paradigm for 3d point cloud
  completion.
\newblock In \emph{Proc.~of the Int.~Conf.~on Learning Representations (ICLR)},
  2022.

\bibitem[Rold{\~a}o et~al.(2020)Rold{\~a}o, de~Charette, and
  Verroust-Blondet]{roldao2020threedv}
Luis Rold{\~a}o, Raoul de Charette, and Anne Verroust-Blondet.
\newblock {LMSCNet: Lightweight Multiscale 3D Semantic Completion}.
\newblock In \emph{Proc.~of the Intl.~Conf.~on 3D Vision (3DV)}, 2020.

\bibitem[Vizzo et~al.(2022)Vizzo, Mersch, Marcuzzi, Wiesmann, Behley, and
  Stachniss]{vizzo2022ral}
Ignacio Vizzo, Benedikt Mersch, Rodrigo Marcuzzi, Louis Wiesmann, Jens Behley,
  and Cyrill Stachniss.
\newblock Make it dense: Self-supervised geometric scan completion of sparse 3d
  lidar scans in large outdoor environments.
\newblock \emph{IEEE Robotics and Automation Letters (RA-L)}, 7\penalty0
  (3):\penalty0 8534--8541, 2022.

\bibitem[Zhou et~al.(2021)Zhou, Du, and Wu]{zhou2021iccv-sgac}
Linqi Zhou, Yilun Du, and Jiajun Wu.
\newblock {3D Shape Generation and Completion Through Point-Voxel Diffusion}.
\newblock In \emph{Proc.~of the IEEE/CVF Intl.~Conf.~on Computer Vision
  (ICCV)}, 2021.

\end{thebibliography}
